\documentclass[11pt,letterpaper]{article}

\usepackage{xcolor}
\definecolor{navy}{rgb}{0,0.1,0.4}
\definecolor{navy-cap}{rgb}{0,0.1,0.5}
\definecolor{lgray}{gray}{0.90}
\definecolor{dkgreen}{rgb}{0,0.6,0}
\definecolor{gray}{rgb}{0.5,0.5,0.5}
\definecolor{mauve}{rgb}{0.58,0,0.82}

\usepackage{sectsty}
\sectionfont{\fontsize{12}{15}\selectfont \color{navy}}
\subsectionfont{\fontsize{12}{15}\selectfont \color{navy}}
\subsubsectionfont{\fontsize{12}{15}\selectfont \color{navy}}

\usepackage[margin = 1 in]{geometry}

\usepackage{times}
\usepackage[font={footnotesize}, labelfont={bf}, labelsep=space, justification=justified, skip=0pt]{caption}

\usepackage{titlesec}
\usepackage{tikz}
\titlespacing{\section}{0pt}{0.75em}{0.25em}
\titlespacing{\subsection}{0pt}{0.5em}{0.25em}
\titlespacing{\subsubsection}{0pt}{0.5em}{0.25em}

\usepackage{array,longtable}
\usepackage{xltabular}
\usepackage{multirow}
\usepackage{pifont}
\newcommand{\xmark}{\ding{55}}%

\usepackage[font={color=navy-cap, footnotesize}, labelfont={bf}, labelsep=space, justification=justified, skip=3pt]{caption}
\usepackage{subcaption}
\usepackage{booktabs} 
\usepackage{graphicx}
\usepackage{wrapfig}
\usepackage{enumitem}
\usepackage{hyperref} 
\usepackage{textcomp}
\usepackage{url}
\usepackage{amssymb}
\usepackage{bm}
\usepackage{amsmath}
\DeclareMathOperator*{\argmax}{argmax}
\DeclareMathOperator*{\argmin}{argmin}
\usepackage{dirtytalk}
\usepackage{pdfpages} 
\usepackage{soul} 
\usepackage{fancyhdr}
\usetikzlibrary{patterns}
\usepackage{pgfplots}
\usepackage{floatrow}
\usepackage{setspace}
\usepackage{bbm}

\usepackage[
backend=biber,
style=numeric-comp,
sorting=none,
maxnames = 10
]{biblatex}
\usepackage{cleveref} 

\newcommand{\cmt}[1]{} 
\newcommand{\ie}{i.e., }
\newcommand{\eg}{e.g., }

\newcommand{\RR}{\ensuremath{\mathbb{R}}}

\newcommand{\name}{Probabilistic neural data fusion}
\newcommand{\acronym}{Pro-NDF }
\newcommand{\acronymNoSpace}{Pro-NDF}
\newcommand{\indsamp}[2]{#1^{(#2)}} 
\newcommand{\expectation}{\mathop{\mathbb{E}}}
\newcommand{\cov}{\text{cov}}
\newcommand{\xb}{\boldsymbol{x}}
\newcommand{\xbi}{\indsamp{\boldsymbol{x}}{i}}
\newcommand{\xbj}{\indsamp{\boldsymbol{x}}{j}}
\newcommand{\xbst}{\boldsymbol{x}^{*}}

\newcommand{\yb}{\boldsymbol{y}}
\newcommand{\yhati}{\indsamp{\hat{y}}{i}}
\newcommand{\zb}{\boldsymbol{z}}

\newcommand{\tb}{\boldsymbol{t}}
\newcommand{\tbi}{\indsamp{\boldsymbol{t}}{i}}
\newcommand{\tbj}{\indsamp{\boldsymbol{t}}{j}}
\newcommand{\tbst}{\boldsymbol{t}^{*}}

\newcommand{\pb}{\boldsymbol{p}}
\newcommand{\pbst}{\boldsymbol{p}^{*}}
\newcommand{\pbpr}{\boldsymbol{p}^{'}}
\newcommand{\vb}{\boldsymbol{v}}
\newcommand{\Ab}{\boldsymbol{A}}
\newcommand{\rb}{\boldsymbol{r}}
\newcommand{\Rb}{\boldsymbol{R}}
\newcommand{\oneb}{\boldsymbol{1}}

\newcommand{\omegab}{\boldsymbol{\omega}}
\newcommand{\Omegab}{\boldsymbol{\Omega}}
\newcommand{\thetab}{\boldsymbol{\theta}}

\newcommand{\varphib}{\boldsymbol{\varphi}}

\newcommand{\zetab}{\boldsymbol{\zeta}}
\newcommand{\meanhat}{\hat{\mu}}
\newcommand{\stddevhat}{\hat{\sigma}}
\newcommand{\meanhati}{\indsamp{\meanhat}{i}} 
\newcommand{\stddevhati}{\indsamp{\stddevhat}{i}}
\newcommand{\combinp}{\boldsymbol{u}} 
\newcommand\combinps[1]{\boldsymbol{u}^{s_{#1}}}
\newcommand{\BIIIinp}{\boldsymbol{\nu}} 

\newcommand{\meanout}{\hat{\mu}(\combinp)}
\newcommand{\meansqout}{\hat{\mu}^{2}(\combinp)}
\newcommand{\stdout}{\hat{\sigma}(\combinp)}

\newcommand{\indmeanout}{\hat{\mu}_{\thetab_{j}}(\combinp)}
\newcommand{\indmeansqout}{\hat{\mu}^{2}_{\thetab_{j}}(\combinp)}
\newcommand{\indstdout}{\hat{\sigma}_{\thetab_{j}}(\combinp)}
\newcommand{\indstdsqout}{\hat{\sigma}^{2}_{\thetab_{j}}(\combinp)}

\newcommand{\tbc}{\boldsymbol{t}^{c}}
\newcommand{\ts}{t^{s}}
\newcommand{\tc}{t^{c}}
\newcommand{\Abs}{\boldsymbol{A}^{s}}
\newcommand{\Abc}{\boldsymbol{A}^{c}}
\newcommand{\zbs}{\boldsymbol{z}^{s}}
\newcommand{\zbc}{\boldsymbol{z}^{c}}
\newcommand{\zs}{z^{s}}
\newcommand{\zc}{z^{c}}
\newcommand{\ub}{\hat{{{u}}}} 
\newcommand{\lb}{\hat{{l}}}
\newcommand{\ubi}{\indsamp{\ub}{i}}
\newcommand{\lbi}{\indsamp{\lb}{i}}
\newcommand{\yi}{\indsamp{y}{i}}
\newcommand{\loss}{\mathcal{L}}
\newcommand{\LMGPsig}{s}
\newcommand{\LMGPsighat}{\hat{s}}
\newcommand{\LMGPmu}{m}
\newcommand{\LMGPmuhat}{\hat{m}}
\newcommand{\yhb}{\boldsymbol{y}^{h}}
\newcommand\ylb[1]{\boldsymbol{y}^{l_{#1}}}

\newcommand{\xhb}{\boldsymbol{x}^{h}}
\newcommand\xlb[1]{\boldsymbol{x}^{l_{#1}}}
\newcommand\xsb[1]{\boldsymbol{x}^{s_{#1}}}
\newcommand{\yh}{y^{h}}
\newcommand\yl[1]{y^{l_{#1}}}
\newcommand\ys[1]{y^{s_{#1}}}
\newcommand\yshat[1]{\hat{y}^{s_{#1}}}
\newcommand\fshat[1]{\hat{f}^{s_{#1}}}

\newcommand{\nh}{n^{h}}
\newcommand\nl[1]{n^{l_{#1}}}
\newcommand\ns[1]{n^{s_{#1}}}
\newcommand{\lf}{LF } 
\newcommand{\hf}{HF }
\newcommand\norm[1]{\left\lVert#1\right\rVert}
\newcommand\brackets[1]{\left[#1\right]}
\newcommand\parens[1]{\left(#1\right)}
\newcommand\braces[1]{\left\{#1\right\}}
\newcommand\bars[1]{\left|#1\right|}
\newcommand\redt[1]{\textcolor{red}{#1}} 
\newcommand{\yhAna}{$\frac{1}{0.1 x^3+x^2+x+1}$}
\newcommand{\ylIAna}{$\frac{1}{\redt{0.2} x^3+x^2+x+1}$}
\newcommand{\ylIIAna}{$\frac{1}{\redt{0}\times x^3+x^2+x+1}$}
\newcommand{\ylIIIAna}{$\frac{1}{\redt{0}\times x^3+x^2+\redt{0}\times x+1}$}

\newcommand{\yhWWI}{$ 0.036 S_\omega^{0.758} W_{f \omega}^{0.0035}\left(\frac{A}{\cos ^2(\Lambda)}\right)^{0.6} q^{0.006}\times$} \newcommand{\yhWWII}{$\lambda^{0.04}\left(\frac{100 t_c}{\cos (\Lambda)}\right)^{-0.3} + \left(N_z W_{d g}\right)^{0.49}+S_\omega W_p $}
\newcommand{\ylIWWI}{$ 0.036 S_\omega^{0.758} W_{f \omega}^{0.0035}\left(\frac{A}{\cos ^2(\Lambda)}\right)^{0.6} q^{0.006}\times$} \newcommand{\ylIWWII}{$\lambda^{0.04}\left(\frac{100 t_c}{\cos (\Lambda)}\right)^{-0.3} + \left(N_z W_{d g}\right)^{0.49}+\redt{1} \times W_p $}
\newcommand{\ylIIWWI}{$ 0.036 S_\omega^{\redt{0.8}} W_{f \omega}^{0.0035}\left(\frac{A}{\cos ^2(\Lambda)}\right)^{0.6} q^{0.006}\times$} \newcommand{\ylIIWWII}{$\lambda^{0.04}\left(\frac{100 t_c}{\cos (\Lambda)}\right)^{-0.3} + \left(N_z W_{d g}\right)^{0.49}+\redt{1} \times W_p $}
\newcommand{\ylIIIWWI}{$ 0.036 S_\omega^{\redt{0.9}} W_{f \omega}^{0.0035}\left(\frac{A}{\cos ^2(\Lambda)}\right)^{0.6} q^{0.006}\times$} \newcommand{\ylIIIWWII}{$\lambda^{0.04}\left(\frac{100 t_c}{\cos (\Lambda)}\right)^{-0.3} + \left(N_z W_{d g}\right)^{0.49}+\redt{0} \times W_p $}

\newcommand{\yhBH}{$\frac{2\pi T_{u} \parens{H_{u} - H_{l}}}
{\ln{\parens{\frac{r}{r_w}}}
\parens{1 +
\frac{2 L T_{u}}{\ln{\parens{\frac{r}{r_w}}} r_{w}^{2} k_{w}} +
\frac{T_u}{T_l}}}$}
\newcommand{\ylIBH}{$\frac{2\pi T_{u} \parens{H_{u} - \redt{0.8} H_{l}}}
{\ln{\parens{\frac{r}{r_w}}}
\parens{1 +
\frac{\redt{1} L T_{u}}{\ln{\parens{\frac{r}{r_w}}} r_{w}^{2} k_{w}} +
\frac{T_u}{T_l}}}$}
\newcommand{\ylIIBH}{$\frac{2\pi T_{u} \parens{H_{u} - \redt{3} H_{l}}}
{\ln{\parens{\frac{r}{r_w}}}
\parens{1 +
\frac{\redt{8} L T_{u}}{\ln{\parens{\frac{r}{r_w}}} r_{w}^{2} k_{w}} + \redt{0.75}
\frac{T_u}{T_l}}}$}
\newcommand{\ylIIIBH}{$\frac{2\pi T_{u} \parens{\redt{1.1} H_{u} - H_{l}}}
{\ln{\parens{\frac{\redt{4} r}{r_w}}}
\parens{1 +
\frac{\redt{3} L T_{u}}{\ln{\parens{\frac{r}{r_w}}} r_{w}^{2} k_{w}} +
\frac{T_u}{T_l}}}$}
\newcommand{\ylIVBH}{$\frac{2\pi T_{u} \parens{\redt{1.05} H_{u} - H_{l}}}
{\ln{\parens{\frac{\redt{2} r}{r_w}}}
\parens{1 +
\frac{2 L T_{u}}{\ln{\parens{\frac{r}{r_w}}} r_{w}^{2} k_{w}} +
\frac{T_u}{T_l}}}$}
\newcommand\tworow[1]{\multirow{2}{*}{#1}}
\newcommand\Tstrut{\rule{0pt}{2.6ex}}       
\newcommand\Bstrut{\rule[-1.2ex]{0pt}{0pt}} 
\newcommand\TstrutL{\rule{0pt}{3.9ex}}       
\newcommand\BstrutL{\rule[-1.8ex]{0pt}{0pt}} 

    \makeatletter
\def\@fnsymbol#1{\ensuremath{\ifcase#1\or \dagger\or *\or \ddagger\or
   \mathsection\or \mathparagraph\or \|\or **\or \dagger\dagger
   \or \ddagger\ddagger \else\@ctrerr\fi}}
    \makeatother

\title{Probabilistic Neural Data Fusion for Learning from an Arbitrary Number of Multi-fidelity Data Sets}
\date{\vspace{-5ex}}
\usepackage{authblk}
\author[1]{Carlos Mora$^\dagger$}
\author[1]{Jonathan Tammer Eweis-Labolle\thanks{\noindent Equal Contribution.}}
\author[1]{Tyler Johnson}
\author[2]{Likith Gadde}
\author[1]{Ramin Bostanabad\thanks{\noindent Corresponding Author: Raminb@uci.edu \\\href{https://gitlab.com/TammerUCI/pro-ndf}{GitLab repository: https://gitlab.com/TammerUCI/pro-ndf}}}
\affil[1]{Department of Mechanical and Aerospace Engineering, University of California, Irvine}
\affil[2]{Northwood High School, Irvine}

\begin{document}
    \pagenumbering{arabic}
    \maketitle

    \sloppy 
    \noindent \textcolor{navy}{\textbf{Abstract}}

In many applications in engineering and sciences analysts have simultaneous access to multiple data sources. In such cases, the overall cost of acquiring information can be reduced via data fusion or multi-fidelity (MF) modeling where one leverages inexpensive low-fidelity (LF) sources to reduce the reliance on expensive high-fidelity (HF) data. 
In this paper, we employ neural networks (NNs) for data fusion in scenarios where data is very scarce and obtained from an arbitrary number of sources with varying levels of fidelity and cost.
We introduce a unique NN architecture that converts MF modeling into a nonlinear manifold learning problem. Our NN architecture inversely learns non-trivial (e.g., non-additive and non-hierarchical) biases of the LF sources in an interpretable and visualizable manifold where each data source is encoded via a low-dimensional distribution. This probabilistic manifold quantifies model form uncertainties such that LF sources with small bias are encoded close to the HF source. 
Additionally, we endow the output of our NN with a parametric distribution not only to quantify aleatoric uncertainties, but also to reformulate the network's loss function based on strictly proper scoring rules which improve robustness and accuracy on unseen HF data. 
Through a set of analytic and engineering examples, we demonstrate that our approach provides a high predictive power while quantifying various sources uncertainties. Our codes and examples can be accessed via \href{https://gitlab.com/TammerUCI/pro-ndf}{GitLab}. 

\noindent \textcolor{navy}{\textbf{Keywords:}} 
Multi-fidelity Modeling; Uncertainty Quantification; Bayesian Neural Networks; Inverse Problems; Manifold Learning; Data Fusion. 

    \section{Introduction} \label{sec: intro}
In an increasing number of applications in engineering and sciences analysts have simultaneous access to multiple sources of information. For instance, materials' properties can be estimated via multiple techniques such as (in decreasing order of cost and accuracy/fidelity) experiments, direct numerical simulations (DNS), a host of physics-based reduced order models (ROMs), or analytical methods \cite{pilania2017multi,deng2022data,liu2022simple}. In such applications, the overall cost of gathering information about the system of interest can be reduced via \textit{multi-fidelity} (MF) \textit{modeling} or \textit{data fusion} where one leverages inexpensive low-fidelity (LF) sources to reduce the reliance on expensive high-fidelity (HF) data sources. 
In this paper, we employ neural networks (NNs) for MF modeling in scenarios where data is scarce and obtained from multiple sources with varying levels of fidelity and cost (i.e., data is unbalanced since more samples are available from cheaper sources). In particular, our contributions are as follows $(1)$ we introduce a unique NN architecture that not only facilitates data fusion, but also quantifies and visualizes the discrepancies/similarities between all data sources, and $(2)$ we illustrate that a Bayesian treatment, besides alleviating overfitting and providing a probabilistic surrogate (i.e., an emulator), provides the means to develop a novel loss function (based on proper scoring rules) that improves the performance and robustness of the resulting MF NN emulator.


Over the past few decades, many techniques have been developed for building MF surrogates which are used in outer-loop applications such as design optimization \cite{chakraborty2017surrogate, korondi2021multi}, calibration of computer models \cite{absi2016multi}, or Bayesian optimization \cite{zanjani2022multi}. The main motivation behind these techniques is to leverage the correlations between LF and HF data sources (and the fact that sampling from the former is typically cheaper) to improve the predictive performance of the surrogate while reducing the overall data acquisition costs. 
Early works in this field focused primarily on hierarchically linking bi-fidelity data. 
For instance, in space mapping \cite{tao2019input, koziel2008space, bandler1994space} or multi-level \cite{amrit2020fast, koziel2013multi, leifsson2015aerodynamic} techniques the inputs of the LF data are mapped following formulations such as $\xlb{}=F(\xhb{} )$ where $\xlb{}$ and $\xhb{}$ are the inputs of LF and HF sources, respectively. In this equation, $F(\cdot)$ is a transformation function whose predefined functional form is calibrated such that $\yl{}(F(\xhb{}))$ approximates $\yh(\xhb{})$ as closely as possible. These techniques are useful in applications where higher fidelity data are obtained by successively refining the discretization in simulations \cite{amrit2020fast,koziel2013multi}, e.g., by refining the mesh when modeling the flow around an airfoil or estimating the fracture toughness of a microstructure. The main disadvantages of space mapping techniques are that $(1)$ they rely on iterative and time-consuming analysis for choosing a near-optimal functional form for $F(\cdot)$, $(2)$ they cannot jointly fuse more than two data sources at a time, $(3)$ they quantify similarity/discrepancy between the sources based on pre-defined functions whose space may not include the true discrepancy, and $(4)$ they do not quantify some uncertainty sources (such as lack of data) and are rarely formulated within a Bayesian setting that leverages prior information.

A well-known hierarchical bi-fidelity modeling framework is that of Kennedy and O'Hagan (KOH) \cite{kennedy2001bayesian} who assume that the discrepancy between the LF and HF sources is additive (multiplicative terms have also been explored \cite{mcfarland2008multivariate}) and that both sources as well as the discrepancy between them can be modeled via Gaussian processes (GPs). Upon this modeling assumption, KOH find the joint posterior of GPs' hyperparameters via either fully \cite{plumlee2017bayesian, higdon2004combining} or modular Bayesian inference \cite{apley2006understanding, bayarri2007framework, arendt2012improving, arendt2012quantification}.
While KOH's approach considers multiple uncertainties and has been successfully applied to a broad range of applications \cite{stainforth2005uncertainty, zhang2019numerical, gramacy2015calibrating}, it has three main limitations: $(1)$ it only accommodates two data sources at a time, $(2)$ it places a priori independence assumption between the GPs, and $(3)$ it does not provide a low-dimensional, visualizable, and interpretable metric that quantifies the correlations between the data sources. 

Recent works have acknowledged the limitations of hierarchical methods and devised new methodologies to address them. For instance, MF modeling can be achieved via a recursive scheme \cite{jofre2018multi} where a bi-fidelity method is repeatedly applied from the lowest to the highest fidelities. However, such recursive schemes inherit the limitations of bi-fidelity methods, cannot \textit{jointly} fuse multi-source data sets, and are sensitive to the ordering (i.e., the relative accuracy of all sources must be known a priori).

As another example, \cite{gorodetsky2020mfnets} presents MF networks (MFNets): an approach based on directed acyclic graphs that builds a MF surrogate using an arbitrary number of data sources. MFNets accommodate noisy data and are trained via gradient-based minimization of a nonlinear least squares objective. While MFNets can learn non-hierarchical relations between data sources, they: $(1)$ rely on having prior knowledge on a set of latent variables that explain the relations between the sources, $(2)$ assume each source can be surrogated via a linear subspace model, $(3)$ are not probabilistic and also require regularization, $(4)$ impose independence assumption among the data sources to derive the likelihood (i.e., the objective) function, and $(5)$ rely on iterative approaches for finding the optimal graph structure.

Other notable works that have studied the limitations of hierarchical techniques include \cite{morrison2018representing,morrison2019embedded,portone2017stochastic} which are focused on identifying (and correcting) non-additive discrepancies between LF and HF sources. However, the proposed solution in these works is intrusive and relies on some rather strong modeling assumptions that largely limit the applications. These limitations arise because the formulation of the discrepancy is learned via an embedded operator whose functional form and interaction with the LF source are constructed a priori. 

We have recently developed a GP-based approach \cite{eweis2022data} that addresses the above issues by converting MF modeling into a manifold learning problem where the relations between the sources are automatically quantified via an appropriately learnt distance measure. The conversion is achieved via latent map Gaussian processes \cite{eweis2022data} (LMGPs, see \Cref{sec: back-LMGP}) which enable GPs to handle categorical variables and, correspondingly, data fusion: by augmenting the inputs via a categorical variable (which indicates the source of a data point) and then concatenating all the data sets, LMGPs can simultaneously learn from an arbitrary number of information sources. We have shown \cite{eweis2022data} that LMGP-based MF modeling consistently outperforms KOH' approach and can also handle calibration problems.

Following the success of LMGPs in data fusion, in this work we examine the potentials of NNs in matching (and, hopefully, improving) LMGPs' efficiency in MF modeling. Our current studies are motivated by the facts that $(1)$ when viewed as (probabilistic or deterministic) graphical models \cite{RN1070}, NNs provide unique opportunities to use MF data sets to uncover complex hidden relations between the corresponding sources, $(2)$ the recent hardware and software advancements have dramatically accelerated architecture design and training of NNs, and $(3)$ NNs scale to higher dimensions and big data significantly better than GPs. 

Over the past few years some NN-based approaches have been developed for MF modeling \cite{meng2020composite, gorodetsky2020mfnets, de2020transfer, Pawar2022concatenatedNN}. However, most of these works design the network architecture primarily based on hierarchical methods and consequently inherit their limitations. For instance, \cite{meng2020composite} builds two sequentially connected deterministic networks based on KOH's method where the first and second NNs are tasked to emulate the LF and HF sources, respectively. In addition to sharing the limitations of KOH's method, such a sequential bi-fidelity NN requires the LF and HF training data to be available at the same inputs (unless the two parts of the network are trained separately) and also relies on manual tuning of the architecture and loss function. It has been argued \cite{de2020transfer} that such sequentially trained NNs bridge MF modeling with transfer learning where the knowledge gained from the LF data is used in building the NN module that surrogates the HF source.

Non-sequential NNs are rarely used for MF modeling (esp. with $>2$ sources) due to the fact that searching for the optimum architecture (and effectively training it with small data) is a difficult task. We address this challenge by drawing inspiration from LMGPs where we design the architecture such that any number of MF data sets can be simultaneously fused and the overall discrepancies between sources are quantified with visualizable metrics. We also illustrate that making specific parts of the network probabilistic, in addition to being superior to both deterministic and all-probabilistic NNs, enables us to infuse a proper scoring rule \cite{gneiting2007strictly} into the loss function and, in turn, improve the performance of the MF emulator. The particular rule that we adopt is interval score which is frequently used in testing the quality of probabilistic predictions but, to the best of our knowledge, has never been used in the training stage of a probabilistic NN. 
In summary, our major contributions are as follows:
\begin{itemize}
    \item We introduce a unique NN architecture for MF modeling that can fuse an arbitrary number of data sets and quantify both epistemic and aleatoric uncertainties.
    \item We inversely learn the accuracy of the LF sources (with respect to the HF source) and visualize the learned relations in an interpretable manifold.
    \item We show that a probabilistic setting allows us to develop a novel loss function (based on proper scoring rules) that improves the performance of the emulator.
    \item We validate the performance of our approach on analytical and real-world examples and show that it performs on par with the state of the art while providing improved scalability to high dimensions and big data.
\end{itemize}

The rest of this paper is organized as follows. We review the relevant technical background in \Cref{sec: back} and then introduce our approach in \Cref{sec: meth}. We test the performance of our approach on a host of analytical problems and real-world data sets in \Cref{sec: results} and conclude the paper in \Cref{sec: conclusion}.

    \section{Technical Preliminaries} \label{sec: back}
In this section we first review LMGPs which are extensions of GPs that handle categorical inputs and, thus, can readily fuse any number of data sets. Then, we provide some background on Bayesian neural networks (BNNs) which form the foundation of our neural data fusion framework. 

\subsection{Latent Map Gaussian Processes (LMGPs)} \label{sec: back-LMGP}



Let us denote the output and inputs in the training data by $y \in \mathcal{Y} \equiv \RR$ and $\xb = [x_1, x_2, \dots, x_{dx}]^T \in \RR^{dx}$, respectively, with an individual training point $i=1, \dots, n$ denoted by the pair $(\yi, \xbi)$. Assume the training data is a realization from a constant-mean\footnote{GPs (and LMGPs) can also be formulated by using a linear combination of basis functions in place of the constant mean. This formulation relies on prior knowledge of the functional form of the output and can improve performance in extrapolation, see \cite{eweis2022data}.} GP and that the following relation holds:
\begin{equation}
    y(\xb) = \LMGPmu + \xi(\xb)
    \vspace{-0.2cm}
    \label{eqn:GPmodel}
\end{equation}
\noindent where $\LMGPmu$ is the unknown constant mean and $\xi(\xb)$ is a zero-mean GP whose covariance function or kernel is:
\begin{equation}
    \cov (\xi(\xb), \xi(\xb^{'})) = c(\xb,\xb^{'}) = \LMGPsig^2 r(\xb, \xb^{'} )
    \vspace{-0.2cm}
    \label{eqn:GPcovfn}
\end{equation}
\noindent where $\LMGPsig^2$ is the variance of the process and $r(\cdot, \cdot)$ is a parametric correlation function such as the Gaussian:
\begin{equation}
    r(\xb, \xb^{'}) = \exp{\parens{-\sum^{dx}_{i=1} 10^{\omega_i}(x_i - x_i^{'})^2}} = 
    \exp{\parens{-\parens{\xb-\xb^{'}}^{T} 10^{\Omegab} \parens{\xb - \xb^{'}}}}
    \vspace{-0.2cm}
    \label{eqn:GPcorrfn}
\end{equation}
\noindent where $\omegab = [\omega_1, \dots, \omega_{dx}]^T$ are the roughness or scale parameters and $\Omegab = diag(\omegab)$.

The training process and prediction formulas for a GP depend on the choice of the correlation function, which relies on a weighted Cartesian distance metric between any two inputs, see \Cref{eqn:GPcorrfn}. As we recently motivated in \cite{oune2021latent}, to directly use GPs for mixed-variable modeling we reformulate $r(\cdot, \cdot)$ as detailed below such that it can handle categorical (qualitative) inputs. 


Let us denote the categorical inputs by $\tb = [t_1, \dots, t_{dt}]^T$ where the total number of distinct levels for qualitative variable $t_i$ is $\tau_{i}$. To handle mixed inputs, LMGP learns a parametric function that maps categorical variables to some points in a quantitative manifold or latent space\footnote{Multiple mapping functions can also be used to build multiple manifolds. We leverage this in \Cref{sec: results} where we build two manifolds for data fusion problems with categorical or mixed inputs.}. These points (and hence the mapping function) can be incorporated into any standard correlation function, such as the Gaussian, which is reformulated as follows for mixed inputs:
\begin{equation}
    r\left( (\xb, \tb), (\xb^{'}, \tb^{'}) \right) = \exp{\parens{
        -\norm{\zb(\tb) - \zb(\tb^{'})}^2_2 - \parens{\xb - \xb^{'}}^{T} 10^{\Omegab} \parens{\xb - \xb^{'}}
        }}
    \vspace{-0.2cm}
    \label{eqn:LMGPcorrfn}
\end{equation}
\noindent or, equivalently,
\begin{equation}
    r\left( (\xb, \tb), (\xb^{'}, \tb^{'}) \right) = \exp{\parens{-\sum^{dx}_{i=1} 10^{\omega_i}(x_i - x_i^{'})^2}} \times
    \exp{\parens{-\sum^{dz}_{i=1} (z_{i}(\tb) - z_{i}(\tb^{'}))^2}}
    \vspace{-0.2cm}
    \label{eqn:LMGPcorrfn2}
\end{equation}
\noindent where $\norm{\cdot}_2$ denotes the Euclidean 2-norm and $\zb(\tb)=[z_{1}(\tb), \dots, z_{dz}(\tb)]_{1\times dz}$ is the to-be-learned latent space point corresponding to the particular combination of categorical variables denoted by $\tb$. To find these points in the latent space, LMGP assigns a unique vector (i.e., a prior representation) to each combination of categorical variables. Then, it uses matrix multiplication\footnote{More complex transformations based on, \eg NNs, may also be used, although we do not do so in this paper.} to map each of these vectors to a point in a latent space of dimension $dz$:
\begin{equation}
    \zb(\tb) = \zetab(\tb)\Ab
    \vspace{-0.2cm}
    \label{eqn:LMGPlatentmap}
\end{equation}
\noindent where $\zetab(\tb)$ is the $1\times \sum^{dt}_{i=1} \tau_{i}$ unique prior vector representation of $\tb$ and $\Ab$ is a $\sum^{dt}_{i=1} \tau_{i} \times dz$ matrix that maps $\zetab(\tb)$ to $\zb(\tb)$. 
In this paper, we use $d{z} = 2$ since it simplifies visualization and has been shown to provide sufficient flexibility for learning the latent relations \cite{oune2021latent}. 
We construct $\zetab$ via a form of one-hot encoding where we first construct the $1\times \tau_{i}$ vector $\vb^{i} = \left[ v^{i}_{1}, v^{i}_{2}, \dots, v^{i}_{\tau_{i}} \right]$ for each categorical variable $t_i$ such that $v^{i}_{j}=1$ when $t_i$ is at level $k=j$ and $v^{i}_{j}=0$ when $t_i$ is at level $k\neq  j$ for $k\in 1, 2, \dots, \tau_{i}$. Then, we set $\zetab(\tb)=[\vb^{1}, \vb^{2}, \dots, \vb^{d_t}]$. For example, for the two categorical variables $t_1$ and $t_2$ with $2$ and $3$ levels, $\zetab(\tb) = [0, 1, 0, 1, 0]$ encodes the combination where both variables are at level $2$.

To train an LMGP, we use maximum likelihood estimation (MLE) to jointly estimate all of its parameters:
\begin{equation}
    \brackets{\LMGPmuhat, \LMGPsighat^2, \hat{\omegab}, \hat{\Ab}} = 
    \argmax_{\LMGPmu, \LMGPsig^{2}, \omegab, \Ab} \hspace{5mm}
    \bars{2\pi \LMGPsig^{2} \Rb}^{-\frac{1}{2}} \times
    \exp{\parens{-\frac{1}{2} (\yb - \oneb\LMGPmu)^{T} (\LMGPsig^{2}\Rb)^{-1} (\yb - \oneb\LMGPmu)}}
    \vspace{-0.2cm}
    \label{eqn:LMGPMLEone}
\end{equation}
\cmt{\noindent or equivalently
\begin{equation}
    \brackets{\LMGPmuhat, \LMGPsighat^2, \hat{\omega}, \hat{\Ab}} = 
    \argmin_{\LMGPmu, \LMGPsig^{2}, \omegab, \Ab} \hspace{5mm}
    \frac{L}{2}\log{\parens{\LMGPsig^2}} + \frac{1}{2}\log{\parens{\bars{\Rb}}} + 
    \frac{1}{2\LMGPsig^2}(\yb - \oneb\LMGPmu)^{T} (\LMGPsig^{2}\Rb)^{-1} (\yb - \oneb\LMGPmu)
    \vspace{-0.2cm}
    \label{eqn:LMGPMLEtwo}
\end{equation}
where $\log{(\cdot)}$ is the natural logarithm, 
}
\noindent where $\bars{\cdot}$ denotes the determinant operator, $\yb = \brackets{y_1, \dots, y_n}^{T}$ is the $n\times 1$ vector of outputs in the training data, $\Rb$ is the $n\times n$ correlation matrix with the $(i, j)^{\text{th}}$ element $R_{ij}=r\parens{(\xbi, \tbi), (\xbj, \tbj)}$ for $i,j=1,\dots,n$, and $\oneb$ is a $n\times1$ vector of ones.

\cmt{Using the method of profiling \cite{RN783}, \Cref{eqn:LMGPMLEtwo} can be simplified to:
\begin{equation}
    \brackets{\hat{\omegab}, \hat{\Ab}} = 
    \argmin_{\omegab, \Ab} \hspace{5mm}
    n \log{\parens{\LMGPsighat^2}} + \log{\parens{\bars{\Rb}}} = 
    \argmin_{\omegab, \Ab} \hspace{5mm} L
    \vspace{-0.2cm}
    \label{eqn:LMGPMLEthree}
\end{equation}
\noindent where $\LMGPsighat^{2} = \frac{1}{L} \parens{\yb - \oneb \LMGPmuhat}^{T} \Rb^{-1} \parens{\yb - \oneb \LMGPmuhat}$ and $\LMGPmuhat = \brackets{\oneb^{T}\Rb^{-1}\oneb}^{-1} \brackets{\oneb^{T}\Rb^{-1}\yb}$. \Cref{eqn:LMGPMLEthree} can be efficiently solved via a gradient-based optimization technique \cite{RN649, RN783}.
}
After estimating the hyperparameters, we use the conditional distribution formulas to predict the response distribution at the arbitrary point $\pbst = (\xbst, \tbst)$. The mean and variance of this normal distribution are:
\begin{equation}
    \expectation \brackets{y \parens{\pbst}} = 
    \LMGPmuhat + \rb^{T} \parens{\pbst} \Rb^{-1}\parens{\yb - \oneb \LMGPmuhat}
    \vspace{-0.2cm}
    \label{eqn:LMGPpredmean}
\end{equation}

\begin{equation}
    \cov \parens{y(\pbst), y(\pbpr)} = \LMGPsighat^{2} r(\pbst, \pbpr) = \LMGPsighat^{2} \braces{1- \rb^{T}(\pbst)\Rb^{-1} \rb(\pbpr) +
    g(\pbst)(\oneb^{T}\Rb^{-1}\oneb)^{-1}g(\pbpr)}
    \vspace{-0.2cm}
    \label{eqn:LMGPpredcov}
\end{equation}
\noindent where $\expectation$ denotes expectation, $\rb \parens{\pbst}$ is an $(n\times 1)$ vector with the $i^{\text{th}}$ element $r\parens{\indsamp{\pb}{i}, \pbst}$, and $g \parens{\pbst}=1 - \oneb^{T}\Rb^{-1}\rb \parens{\pbst}$.

To perform data fusion via LMGP, we re-frame multi-fidelity modeling as a manifold learning problem. Assume that we have $ds$ data sources whose inputs and outputs are denoted by $\xsb{i}, \ys{i}$, respectively, with $i=1,\dots, ds$. We first pre-process the data by appending the inputs with a single categorical variable $\ts$ with $ds$ levels (hereafter referred to as the source index variable) that distinguishes the data sources. Specifically, we add $\ts$ at level $i$ for source $s_i$, \ie $\xsb{i} \rightarrow \brackets{\xsb{i}, \boldsymbol{i}_{\ns{i}\times 1}}$, where $\boldsymbol{i}_{\ns{i}\times 1}$ is an $\ns{i}\times 1$ vector of $i$'s and $\ns{i}$ is the number of data points for source $s_i$. We then combine the data for all sources into one unified data set and fit an LMGP directly it, \ie we fit LMGP to \textit{all} of the data from \textit{all} sources at once. 

The fitted LMGP can provide predictions for any desired data source based on the level used for $\ts$ and as such is an emulator for all of the data sources. Additionally, since the data sources are distinguished via a categorical variable, LMGP learns the correlations between them via a visualizable latent representation and uses these correlations to improve its predictions \cite{eweis2022data}. In the case that the raw inputs contain categorical variables $\tbc$, we use separate mappings for $\ts$ and $\tbc$, \ie we assign unique priors $\zetab(\ts)$ and $\zetab(\tbc)$ which LMGP uses to find mapping matrices $\Abs$ and $\Abc$. The latent points corresponding to each mapping are then $\zbs$ and $\zbc$, respectively.

Note that the correlation function in \Cref{eqn:LMGPcorrfn2} depends directly on the euclidean distance between a pair of latent points. This means that relative distances in the latent space directly correspond to correlations, \eg if a pair of data sources $\ys{1}$ and $\ys{2}$ have corresponding latent points with a distance $\Delta$ in the latent space then this directly implies by \Cref{eqn:LMGPcorrfn2} that LMGP has found those two sources to have a correlation of $\exp{(-\Delta^{2})}$. 

\subsection{Bayesian Neural Networks} \label{sec: back-BNNs}


Feedforward neural networks (FFNNs) are one of the most common models used in deep learning and their main goal is to learn the underlying function $f(\xb)$ that maps the inputs $\xb$ to the target $y$ \cite{lecun2015deep}. To this end, an FFNN defines the mapping $\hat{f}(\xb;\thetab)$ whose parameters $\thetab$ are estimated such that $\hat{y} = \hat{f}(\xb;\thetab)$ best approximates $f(\xb)$. NN-based approaches for MF emulation can provide attractive advantages since they are universal function approximators \cite{hornik1989multilayer} and can handle high-dimensional inputs and large data sets. In this subsection, we first describe the working principle of FFNNs and motivate the use of BNNs and Bayes by backprop \cite{blundell2015weight}. 

FFNNs propagate information from the inputs $\xb$ to the output $y$ through intermediate computations that define $\hat{f}$. They are traditionally built via a succession of $L$ layers where $L-2$ hidden layers are placed between the input and output layers. The output of layer $k$ is denoted by $\zb_k$ and is obtained as follows:
\vspace{-0.2cm}
\begin{align}
    \zb_1 &=\xb, \\
    \zb_k &=\phi_{k}\left(\boldsymbol{W}_{k} \zb_{k-1}+\boldsymbol{b}_{k}\right) \quad \forall \ k \in[2, L-1], \\
    \hat{y} &=\phi_{L}\left(\boldsymbol{W}_{L} \boldsymbol{z}_{L-1}+{b}_{L}\right)
    \vspace{-0.2cm}
    \label{eqn:NN_forwardpass}
\end{align}
where $\phi$ is the (typically non-linear) activation function. The parameters $\thetab_k=(\boldsymbol{W}_k,\boldsymbol{b}_k)$, where $\boldsymbol{W}_k$ and $\boldsymbol{b}_k$ are the weight matrices and bias vectors, respectively, correspond to the connections between the $(k-1)^{th}$ and $k^{th}$ layer. For brevity, we denote the parameters of the entire network by $\thetab$.

From a statistical perspective, an FFNN aims to learn the conditional distribution $P(y|\xb; \thetab)$ given the noisy data set $\mathcal{D}$ with independent and identically distributed samples:
\begin{equation}
    \yi = f(\xbi) + \epsilon \approx \hat{f}(\xbi;\thetab) + \epsilon
    \label{eqn:regression_problem}
\end{equation}
\noindent where $\epsilon \sim \mathcal{N}(0, \sigma^2)$ represents noise. \Cref{eqn:regression_problem} indicates that $P(\yi|\xbi) \sim \mathcal{N}(f(\xbi), \sigma^2)$ and hence the conditional probability $P\left(\mathcal{D} | \thetab\right)$ can be written as:
\begin{align}
    P(\mathcal{D} | \thetab) &= 
    \prod_{i=1}^{n} P(\yi | \xbi,\thetab)P(\xbi) = \prod_{i=1}^{n} \mathcal{N}(\yi;\yhati, \sigma^2)P(\xbi) \nonumber \\
    &= \prod_{i=1}^{n} \frac{1}{\sigma \sqrt{2\pi}} \exp{\left(-\frac{1}{2\sigma^2}\left(\yi-\yhati \right)^2\right)}P(\xbi)
    \vspace{-0.2cm}
    \label{eqn:likelihood}
\end{align}

\noindent Since the likelihood function $L(\thetab) \equiv P(\mathcal{D} | \thetab)$, the parameters $\thetab$ can be estimated by maximizing $L(\thetab)$ (the dependence on $\xb$ is dropped for brevity):
\begin{equation}
    \thetab_{MLE} = \arg \max _{\thetab} L(\thetab) 
    \equiv \arg \min _{\thetab} \hspace{2mm}-\log P(\mathcal{D} | \thetab)  = 
    \arg \min _{\thetab} \frac{1}{n} \sum_{i=1}^{n} \left(\yi-\hat{f}(\xbi;\thetab)\right)^2
    \label{eqn:weights_MLE}
\end{equation}
 \noindent which is equivalent to minimizing the mean squared error (MSE) of the predictions $\hat{y} = \hat{f}(\xb;\thetab)$ with respect to the targets $y$. \Cref{eqn:weights_MLE} can be updated via Bayes rule to consider prior knowledge on $\thetab$ in the optimization. These maximum a posteriori (MAP) estimates are obtained via:
\begin{align}
    \thetab_{MAP} &= \arg \max _{\thetab} P\left(\thetab | \mathcal{D}\right) = \arg \max _{\thetab} \log P\left(\thetab | \mathcal{D}\right) = \arg \max _{\thetab} \log P\left(\mathcal{D} | \thetab\right) + \log P(\thetab)
    \vspace{-0.2cm}
    \label{eqn:weights_MAP}
\end{align}
where the first term recovers MSE as in \Cref{eqn:weights_MLE} and the second term depends on the prior distribution assigned to the parameters. \Cref{eqn:weights_MAP} illustrates that Gaussian and Laplacian priors are equivalent to $L2$ and $L1$ regularization, respectively \cite{lecun2015deep, blundell2015weight}.

FFNNs are likely to overfit in scenarios where data is scarce. Additionally, they cannot directly quantify prediction uncertainty and are often overconfident in extrapolation \cite{guo2017calibration}. BNNs are developed to address these issues \cite{mitros2019validity, kristiadi2020being}. In BNNs, the weights are endowed with probability distributions (rather than single point estimates) which naturally results in probabilistic predictions and can dramatically reduce overfitting via parameter regularization and model averaging. 

Predictions via a BNN requires sampling from the posterior distribution of the parameters, i.e., $P\left(\thetab | \mathcal{D}\right)$, which does not have a closed form and is highly complex. Over the past few years, various techniques have been developed to obtain samples from $P\left(\thetab | \mathcal{D}\right)$ (or an approximation thereof). The most popular techniques are based on either Markov Chain Monte Carlo (MCMC) \cite{hastings1970monte} or variational inference (VI) \cite{blei2017variational} which, unlike MCMC, learns an approximation of the posterior distribution.

Although MCMC methods are arguably the best techniques for sampling from the \textit{exact} posterior, their lack of scalability makes them inefficient for BNNs of any practical size \cite{jospin2022hands}. 
Hence, we employ Bayes by backprop \cite{blundell2015weight} which is a variational method that approximates $P\left(\thetab | \mathcal{D}\right)$ with the parameterized distribution $q(\thetab|\varphib)$ . The parameters $\varphib$ are learned by minimizing the Kullback–Leibler (KL) divergence between the true and approximated posteriors:
\begin{align}
    &\operatorname{KL}[q(\thetab|\varphib)||P(\thetab|\mathcal{D})]=
    \int q(\thetab | \varphib) \log \left(\frac{q(\thetab | \varphib)}{P(\thetab | \mathcal{D})}\right) d \thetab = 
    \int q(\thetab | \varphib) \log \left(\frac{q(\thetab | \varphib)P(\mathcal{D})}{P(\mathcal{D} | \thetab)P(\thetab)}\right) d \thetab \nonumber \\
    &= \int q(\thetab | \varphib) \log P(\mathcal{D}) d \thetab+\int q(\thetab | \varphib) \log \left(\frac{q(\thetab | \varphib)}{P(\thetab)}\right) d \thetab-\int q(\thetab | \varphib) \log P(\mathcal{D} | \thetab) d \thetab \nonumber \\
    &= \log P(\mathcal{D})+\operatorname{KL}[q(\thetab || \varphib) | P(\thetab)]-\mathbb{E}_{q(\thetab|\varphib)}[\log P(\mathcal{D}|\thetab)]
    \vspace{-0.2cm}
    \label{eqn:KLdiv}
\end{align}
\noindent where Bayes rule is applied to $P(\thetab | \mathcal{D})$ in the first line. Then, the parameters $\varphib$ are estimated by minimizing \Cref{eqn:KLdiv}: 
\begin{align}
    \varphib^* = \argmin_{\varphib} \operatorname{KL}[q(\thetab|\varphib)||P(\thetab|\mathcal{D})] = \argmin_{\varphib} \operatorname{KL}[q(\thetab|\varphib)||P(\thetab)]-\mathbb{E}_{q(\thetab|\varphib)}[\log P(\mathcal{D}|\thetab)]
    \vspace{-0.2cm}
    \label{eqn:VI}
\end{align}
where the term $\log P(\mathcal{D})$ is excluded as it is constant. \Cref{eqn:VI} aims to minimize the sum of two terms. The second term corresponds to the expectation of the negative log-likelihood while the first term acts as a regularizer and corresponds to the KL divergence between the approximated posterior and the prior. 
    \section{Probabilistic Neural Data Fusion} \label{sec: meth}

Designing a multi-fidelity NN that leverages an ensemble of LF data sets to better learn an HF source is a very challenging task because of the following major reasons:
\begin{enumerate}
    \item The relations among the data sources can be unknown. For instance, in the Rational example (see \Cref{tab:analytic_functions} in \Cref{sec: app-analytic}) there are three LF sources whose biases are \textit{not} additive. Additionally, these LF sources are not hierarchically ordered in the sense that the second LF source is more accurate than the first one. 
    \item There are typically (but not always) more LF data available since LF sources are generally cheaper compared to the HF source. Learning from such an unbalanced MF data is quite difficult especially in the presence of scarce HF data (as an example, see the sample sizes for the engineering applications described in \Cref{sec: app-real_world}). 
    \item NNs can be built in many ways and, as shown in \Cref{sec: results}, their performance heavily depends on their architecture and training mechanism. Building an optimum\footnote{We measure optimality in terms of NN's error in predicting unseen data from the HF source.} NN with small, unbalanced, and MF data is even more difficult since the sensitivity to the architecture and training mechanism considerably increases. 
\end{enumerate}

We propose to address the above challenges by converting MF modeling to a manifold\footnote{A manifold or a latent-space is a compact representation of a high-dimensional object such as an image.} learning problem which is then solved via an NN. We design the architecture, loss function, and training mechanism of this NN with a particular focus on uncertainty sources that include data scarcity (especially HF samples), noise with unknown variance (which can affect any of the data sources), non-trivial biases of LF sources, and data imbalances. 

\begin{figure*}[!t] 
    \centering
    \includegraphics[page=1, width = 1.0\textwidth]{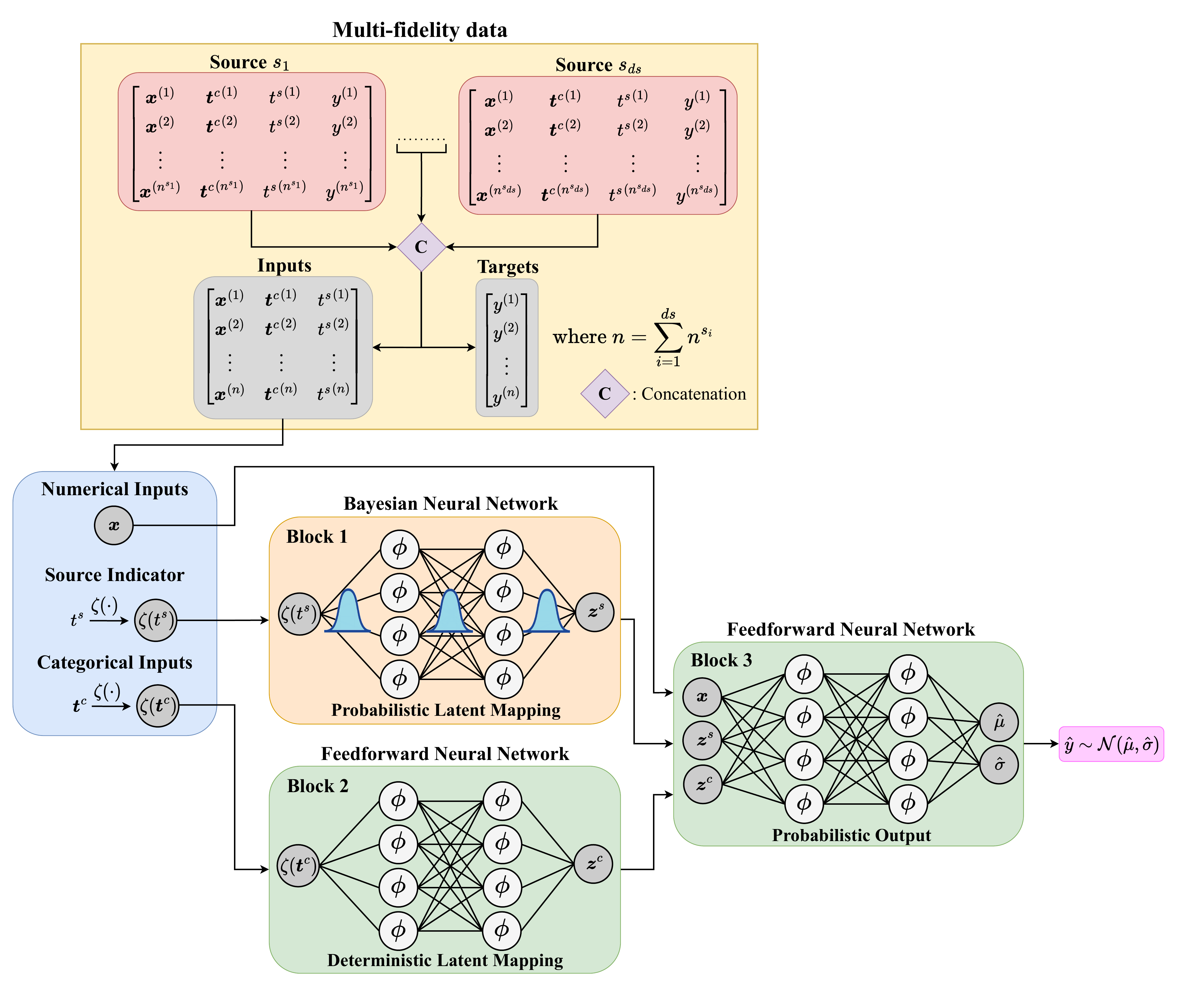}
    \vspace{-0.3cm}
    \caption{\textbf{\name \ (\acronymNoSpace):} The proposed architecture allows to combine an arbitrary number of sources by appending a source indicator variable to the data sets and then concatenating them. \acronym consists of three blocks that perform separate tasks related to MF modeling: (1) Block 1 is a BNN that maps a quantitative prior representation of the source indicator $\zetab(\ts)$ to a continuous manifold, (2) Block 2 is an FFNN that maps a quantitative prior representation of the categorical inputs $\zetab(\tbc)$ to a continuous manifold, and (3) Block 3 is an FFNN with a probabilistic output that maps the numerical inputs and the latent variables to a parametric distribution.}
    \label{fig:NN_architecture}
\end{figure*}

As schematically demonstrated in \Cref{fig:NN_architecture}, we convert MF modeling to manifold learning by augmenting the input space with the categorical variable $\ts$ whose levels (e.g., $\{'1', '2', \cdots\}$ or $\{a, b, \cdots\}$) indicate the source that generates a sample. We then map this \textit{source indicator} variable to a low-dimensional manifold via a BNN (see Block 1 in \Cref{fig:NN_architecture}). If the original input space has the categorical variables $\tbc$, we similarly map them to a manifold (but this time we use a deterministic NN, see Block 2 in \Cref{fig:NN_architecture}). Afterwards, we combine the latent variables of these two manifolds with the quantitative inputs $\xb$ via a deterministic NN, see Block 3 in \Cref{fig:NN_architecture}. As opposed to the other two blocks, we require Block 3 to produce a normal probability distribution in order to capture aleatoric uncertainties. Finally, we train the entire network on the entire\footnote{By entire, we mean the combined data sets from all sources.} data using our custom loss function that noticeably improves the prediction intervals.

In the following subsections, we elaborate on our rationale for designing a multi-block architecture and a custom loss function in \Cref{sec: meth-architecture} and \Cref{sec: meth-loss}, respectively. Then, we provide some details on the training and inference stages in \Cref{sec: meth-trainpredict}. 

\subsection{Multi-Block Architecture} \label{sec: meth-architecture}
Each block of our network is designed to address particular challenges associated with MF modeling. 
Specifically, the BNN of Block 1 maps a quantitative prior representation $\zetab(\ts)$ of the source indicator variable $\ts$ to a continuous manifold $\zbs$. We design $\zetab(\ts)$ by one-hot encoding $\ts$ to merely inform the network about the source that generates a sample\footnote{If there is some prior knowledge about the relation among the sources, $\zetab(\ts)$ can be designed to reflect it. We do not pursue designing such informative priors in this work.}. 
We build $\zbs$ based on a categorical variable because it forces the manifold to uncover the relations between sources (i.e., the levels of $\ts$). These relations are represented as distances in $\zbs$ where sources that produce similar data are encoded with close-by points (see \Cref{sec: results} for multiple examples). This distance learning is in sharp contrast to existing approaches since $(1)$ it does not assume there is any hierarchy between the data sources, 
$(2)$ it is scalable to an arbitrary number of data sets, 
$(3)$ it enables training the entire network via all available samples,
$(4)$ it is visualizable and interpretable which helps in identifying anomalous data sources,  
and $(5)$ it does not assume any specific form (e.g., additive, multiplicative, etc.) for the biases of LF sources.

Block 1 is the only part of our network where the weights and biases are endowed with probability distributions. We make this choice to better learn model form errors and more accurately quantify the epistemic uncertainties due to lack of data and source-wise discrepancies. We note that, while the outputs of Block 1 do \textit{not} parameterize a probability distribution, they are probabilistic by nature since they are obtained by propagating the deterministic vector $\zetab(\ts)$ through some probabilistic hidden layers. 

Block 2 is an FFNN that maps the quantitative prior representation $\zetab(\tbc)$ of the categorical inputs $\tbc$ to the manifold $\zbc$ (Block 2 is omitted if the original inputs are purely quantitative). Similar to Block 1, we design $\zetab(\tbc)$ via one-hot encoding and use deterministic outputs. However, unlike Block 1 we use a deterministic FFNN in Block 2 to map $\zetab(\tbc)$ into $\zbc$. We make this decision to reduce the number of parameters and also because the meaning (and hence effects) of categorical inputs across different sources is typically the same\footnote{Due to severe discrepancies such as large model form errors, the effects of a categorical variable on the response may be quite different across the sources.}. 

We set the manifold dimension to $2$ for both Block 1 and Block 2, i.e., $d\zs = d\zc = 2$. While higher dimensions provide more learning capacities, our results in \Cref{sec: results} and those reported elsewhere \cite{RN366, RN367, RN368, RN369, RN373, RN374} indicate that low-dimensional manifolds are quite powerful in learning highly complex relations. For instance, \cite{RN773} shows that a single latent variable can encode \textit{smiling} in images of human faces which is a high-dimensional and complex feature in the original data space. Additionally, our choice simplifies the visualization of the manifolds and reduces the chances of overfitting since we are primarily interested in scarce data applications. 

Block 3 is also an FFNN that maps the numerical inputs and the latent variables in both manifolds to a parametric distribution which represents the output. Block 3 has deterministic weights and biases since source-wise uncertainties are propagated to it via Block 1. However, we equip Block 3 with a probabilistic output because it:
$(1)$ quantifies aleatoric uncertainties that are inherent to the data sets\footnote{The predicted variance also includes epistemic uncertainties that are propagated from Block 1, see \Cref{sec: meth-trainpredict}}, 
and 
$(2)$ enables designing a multi-task loss that considers the quality of the prediction intervals (detailed in \Cref{sec: meth-loss}). Additionally, Block 3 is responsible for learning the behavior for all data sources simultaneously, which allows it to leverage correlations between sources to augment predictions through a process akin to weight sharing.

\subsection{Uncertainty-Focused Loss} \label{sec: meth-loss}

NNs typically provide overconfident predictions especially when they are trained on small and unbalanced data. As explained in \Cref{sec: meth-architecture}, we aim to address this issue by making Block 1 and the network's final output probabilistic. However, for these measures to work, we must develop an effective optimization\footnote{Recall that we use Bayes by backprop which takes a variational approach towards finding the posteriors, see \Cref{sec: back-BNNs}.} scheme where the loss function appropriately rewards prediction intervals (PIs) that are sufficiently wide (but not too wide) to cover unseen data (especially HF data). To design such a loss function, we draw inspiration from strictly proper scoring rules \cite{gneiting2007strictly} and augment \Cref{eqn:VI} with the negatively oriented interval score. Our loss is defined as:
\begin{align}
    \mathcal{L} = \mathcal{L}_{NLL} + \alpha_1 \mathcal{L}_{KL} + \alpha_2 \mathcal{L}_{IS} + \alpha_3 \mathcal{L}_{2} 
    \label{eqn:loss_function}
\end{align}
\noindent where $\mathcal{L}_{NLL}$ refers to the negative log-likelihood, $\mathcal{L}_{KL}$ is the KL divergence between the prior and the variational posterior distributions on the parameters (only applicable for the BNN from Block 1), $\mathcal{L}_{IS}$ denotes the interval score term, and $\mathcal{L}_{2}$ is $L2$ regularization (only applicable for deterministic NNs, i.e., Block 2 and 3). $\alpha_1$, $\alpha_2$  and $\alpha_3$ are hyperparameters that, respectively, determine the relative strengths of $\mathcal{L}_{KL}$, $\mathcal{L}_{IS}$ and $\mathcal{L}_{2}$ compared to $\mathcal{L}_{NLL}$. 
The four terms in \Cref{eqn:loss_function} are calculated as:
\begin{align}
    & \loss_{NLL} = -\frac{1}{N}\sum^{N}_{i=1} \log \mathcal{N}(\yi;\meanhati,\parens{\stddevhati}^2) \label{eqn:lossNLL}\\
    & \loss_{KL} =  \operatorname{KL}[q(\thetab|\varphib)||P(\thetab)] \label{eqn:lossKL}\\
    & \loss_{IS} = \frac{1}{N}\sum^{N}_{i=1}[(\ubi-\lbi)+\frac{2}{\gamma}(\lbi-\yi) \mathbbm{1}\{\yi<\lbi\}+\frac{2}{\gamma}(\yi-\ubi) \mathbbm{1}\{\yi>\ubi\}] \label{eqn:lossIS}\\
    & \loss_{2} = |\thetab|^2 \label{eqn:lossL2}
\vspace{-0.1cm}
\end{align}
\noindent where $\mathcal{L}_{KL}$ is computed via a Monte Carlo approximation, $N$ is the batch size, and $\mathbbm{1}\{\cdot\}$ denotes the indicator function that returns $1$ if the event in brackets is true and $0$ otherwise. The three terms of \Cref{eqn:loss_function} compose a multi-task loss where: $(1)$ the likelihood term $\mathcal{L}_{NLL}$ penalizes the model if the predicted distribution does not match the target distribution, $(2)$ the KL divergence term $\mathcal{L}_{KL}$ favors variational posteriors that are similar to the assumed prior as per \Cref{eqn:VI}, and $(3)$ the interval score term $\mathcal{L}_{IS}$ rewards narrow PIs while penalizing the model for each observation $\yi$ that lies outside the $(1-\gamma) \times 100\% $ prediction interval that spans the range $[\lbi,\ubi]$ where $\lbi = \meanhati - 1.96\stddevhati$ and $\ubi = \meanhati + 1.96\stddevhati$. In this paper, we use $\gamma=5\%$, thus implying that $\mathcal{L}_{IS}$ is minimized by a distribution whose $95\%$ PI is as tight as possible while containing all the training data.

\subsection{Training and Prediction} \label{sec: meth-trainpredict}

In BNNs, the variational posterior of $\thetab$ is typically defined layer-wise as a multivariate Gaussian with mean $\boldsymbol{\mu} \in \mathbb{R}^{c_k}$ and covariance matrix $\boldsymbol{\Sigma} \in \mathbb{R}^{c_k \times c_k}$, i.e., $\mathcal{N}(\boldsymbol{\mu}, \boldsymbol{\Sigma})$, where $c_k$ is the total number of connections between two consecutive layers. Estimating the full covariance matrix requires learning $\mathcal{O}(c_k^2)$ parameters and is thus computationally prohibitive in most applications \cite{jospin2022hands}. To reduce the costs, some simplifications have been adopted in the literature, such as learning diagonal or block diagonal \cite{ritter2018scalable} covariance matrices. 
However, our approach does not suffer from this computational issue since the only Bayesian part of our network is Block 1 (see \Cref{fig:NN_architecture}) whose size is typically very small (we use one hidden layer with $5$ neurons for all the studies in \Cref{sec: results}). Hence, we estimate a dense covariance matrix between any two layers of Block 1 to improve its uncertainty quantification capacity. 
As for the prior, we use a zero mean Gaussian distribution with diagonal covariance matrix which makes the KL term equivalent to  $L2$ regularization with a rate defined by the standard deviation of the prior distribution \cite{fortunato2018revisiting}. Thus, the standard deviation is a hyperparameter that needs to be tuned specifically to each problem. 

\begin{figure*}[!t] 
    \centering
    \includegraphics[page=1, width = 1.0\textwidth]{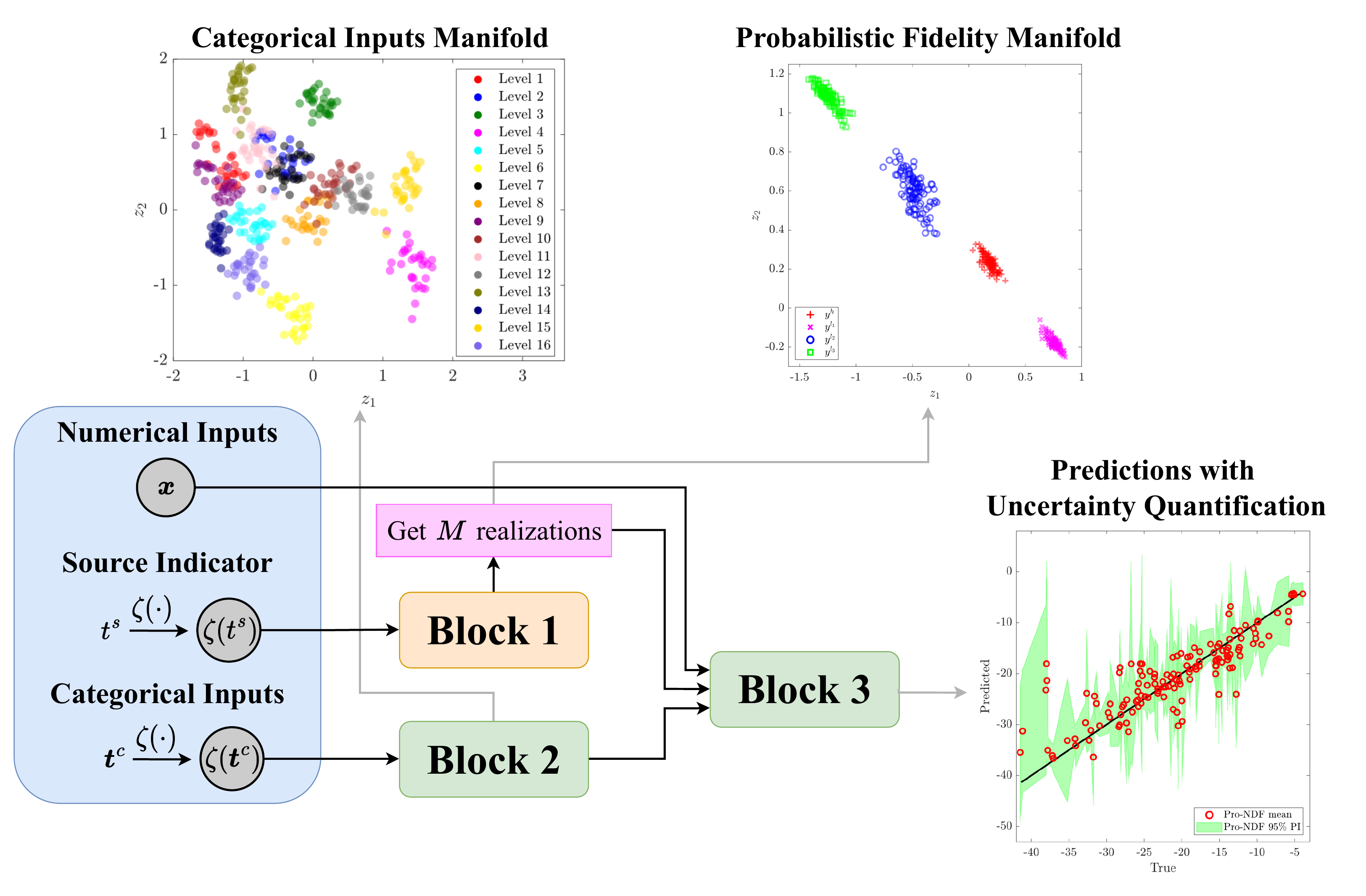}
    \vspace{-0.3cm}
    \caption{\textbf{Outputs of \acronym:} We visualize the outputs of \acronym after it is trained on the MF data of the HOIP data set, which does not have any numerical inputs (see \Cref{sec: app-real_world} for more details). To provide probabilistic predictions that quantify both epistemic as well as aleatoric uncertainties, \acronym learns a probabilistic fidelity manifold (where sources with similar behavior are encoded with close-by distributions) and a deterministic manifold for the categorical inputs.}
    \label{fig:Results_Flowchart}
\end{figure*}

BNNs represent their weights and biases by parameterized distributions which in our case are multivariate normal with dense covariance matrices. In a forward pass during either training or prediction, we take individual samples from these distributions and assign them to the weights and biases. In this way, instead of explicitly obtaining the true posterior distribution of the output of Block 1 (i.e., $\zbs$, see \Cref{fig:NN_architecture}), we obtain an empirical distribution in the $\zbs$ manifold by taking a number of forward passes, see \Cref{fig:Results_Flowchart}. We refer to these forward passes as \textit{realizations} and as explained below we use different number of passes in training versus prediction.

To obtain the response (in training or testing) at the input $\combinp$ using \acronymNoSpace, which contains both a BNN component and a probabilistic output, we use ensemble prediction formulas \cite{Pawar2022concatenatedNN}:
\begin{equation}
    \meanout = \frac{1}{M}\sum_{j=1}^{M} \indmeanout
    \label{eqn:EnsMean}
\end{equation}
\begin{equation}
    \stdout = \frac{1}{M}\sum_{j=1}^{M} \parens{\indstdsqout + \indmeansqout} - \meansqout
    \label{eqn:EnsStd}
\end{equation}
\noindent where $\indmeanout$ and $\indstdout$ are, respectively, the mean and standard deviation of the output distribution in the $j^{th}$ realization and $\thetab_{j}$ are the associated network parameters. 
For predictions with a fitted NN, we use $M=1000$ since it provides a higher accuracy in quantifying the uncertainty associated with learning the fidelity manifold (i.e., $\zbs$). While training the network, we use $M=200$ to reduce the computational costs.

The performance of an NN is highly sensitive to its architecture and hyperparameters if the training data is small, unbalanced, and multi-fidelity. To reduce this sensitivity and leverage the low costs of training a single NN on small data, we perform automated hyperparameter tuning \footnote{We use this approach for all NN-based data fusion approaches (including ours) in \Cref{sec: results}.}. 
To this end, we use RayTune \cite{liaw2018tune} and Hyperopt to find the optimum hyperparameters and architecture by minimizing the five-fold cross-validation errors on predicting the high-fidelity data.

For our approach specifically, we apply the above tuning strategy to the architecture of Block 3, the learning rate of the Adam optimizer, $\alpha_1$, $\alpha_2$ and $\alpha_3$ in \Cref{eqn:loss_function}, the prior standard deviation of weight matrices in Block 1, and the batch size. 
We fix the architectures of Block 1 and Block 2 to one hidden layer with $5$ neurons and the dimension of both manifolds to $2$. The activation function for all the neurons of Block 1 and 3 is hyperbolic tangent, whereas for Block 2 it is the sigmoid function. For more information and full details on implementation, please see \href{https://gitlab.com/TammerUCI/pro-ndf}{our GitLab repository}.

    \section{Results and Discussions}
\label{sec: results}

In this section, we validate our approach on three analytic and two real-world MF problems (detailed in \Cref{sec: app-analytic,sec: app-real_world}) and compare its performance against LMGP and two other existing NN-based approaches which are based on simple feedforward networks or sequential multi-fidelity (SMF) networks which are described in \Cref{sec: app-NN_approaches}. 
The hyperparameters of all the NN-based approaches are tuned as described in \Cref{sec: meth-trainpredict}. We refer the reader to \href{https://gitlab.com/TammerUCI/pro-ndf}{our GitLab repository} for specific details on implementation, estimated hyperparameters, and training/test data. For LMGP, none of its architectural parameters (such as the kernel type, mean function, latent map, etc.) are tuned. 

We first conduct an ablation study in \Cref{sec: results-ablation} to quantify the impacts of our designed architecture, loss function, and probabilistic elements. Then, we test the performance of the four MF approaches on the analytic and real-world problems in \Cref{sec: results-analytic} and \Cref{sec: results-real_world}, respectively. In each problem, the goal is to model the HF source as accurately as possible, \ie to obtain the lowest mean prediction error while maximizing the number of training/test samples that fall in the $95\%$ PI. To this end, we use mean squared-error (MSE) and mean negatively oriented interval score (IS). 
Note that the FFNN and SMF approaches are not probabilistic, \ie they provide point estimates rather than PIs and therefore they are only evaluated based on MSE.

\subsection{Ablation Study} \label{sec: results-ablation}

To evaluate the impact of the key components of \acronymNoSpace, we perform an ablation study on the Rational and DNS-ROM problems which are detailed in \Cref{sec: app-analytic,sec: app-real_world}, respectively. Namely, we analyze the impact of: 
\begin{enumerate}
    \item Using a BNN rather than a deterministic FFNN in Block 1 for probabilistically learning the relations between the data sources. 
    \item Considering $\mathcal{L}_{IS}$ in the loss function of \Cref{eqn:loss_function}.
    \item Fitting the model to the parameters of a distribution instead of a scalar, \ie using a probabilistic output. 
    \item Leveraging the fidelity map to detect the least accurate LF source and, in turn, assessing whether this source helps emulating the HF source. 
\end{enumerate}
Regarding the third item above we note that we no longer use IS in the loss once the probabilistic output is removed. However, we still calculate the IS after training based on the empirical distribution of the fidelity manifold which is produced by the multiple realizations of the BNN component. 

\begin{table}[!b]
    \begin{tabular}{c|c|cccc|ccc|cc}
    \hline
    \multirow{2}{*}{Problem} & \multirow{2}{*}{\begin{tabular}[c]{@{}c@{}}Model\\ Version\end{tabular}} & \multicolumn{4}{c|}{Input data} & \multicolumn{3}{c|}{Components} & \multirow{2}{*}{MSE} & \multirow{2}{*}{IS} \\ \cline{3-9}
     &  & HF & LF1 & LF2 & LF3 & $\mathcal{L}_{IS}$ & PB1 & PO &  &  \\ \hline
    \multirow{5}{*}{Rational} & Base & \checkmark & \checkmark & \checkmark & \checkmark & \checkmark & \checkmark & \checkmark & \bm{$1.65\times 10^{-3}$}  & \bm{$0.22$} \\
    & V1 & \checkmark & \checkmark & \checkmark & \checkmark & \xmark & \checkmark & \checkmark & $3.31\times 10^{-3}$ & $0.26$ \\
    & V2 & \checkmark & \checkmark & \checkmark & \checkmark & \checkmark & \xmark & \checkmark & $2.83\times 10^{-3}$  & $0.24$\\
    & V3 & \checkmark & \checkmark & \checkmark & \checkmark & \xmark & \checkmark & \xmark & $2.01\times 10^{-3}$ & $0.40$ \\
    & V4 & \checkmark & \checkmark & \checkmark & \xmark & \checkmark & \checkmark & \checkmark & $5.68\times 10^{-3}$  & $0.98$ \\ \hline
    \multirow{5}{*}{DNS-ROM} & Base & \checkmark & \checkmark & \checkmark & \checkmark & \checkmark & \checkmark & \checkmark & $8.99\times 10^{9}$  & \bm{$4.84 \times 10^5$} \\
    & V1 & \checkmark & \checkmark & \checkmark & \checkmark & \xmark & \checkmark & \checkmark & $1.23\times 10^{10}$  & $5.89 \times 10^5$ \\
    & V2 & \checkmark & \checkmark & \checkmark & \checkmark & \checkmark & \xmark & \checkmark & $2.09\times 10^{10}$  & $1.11 \times 10^6$ \\
    & V3 & \checkmark & \checkmark & \checkmark & \checkmark & \xmark & \checkmark & \xmark & $1.70\times 10^{10}$  & $4.07 \times 10^6$ \\
    & V4 & \checkmark & \checkmark & \checkmark & \xmark & \checkmark & \checkmark & \checkmark & $\bm{8.17\times 10^{9}}$  & $5.06 \times 10^5$ \\ \hline      
    \end{tabular}
    \caption{\textbf{Results of the ablation study:}
    We evaluate the effect of removing individual components of \acronym from it by reporting the MSE and IS on unseen HF data. All models are trained as discussed in \Cref{sec: meth} (e.g., all models benefit from automatic hyperparameter tuning). For both MSE and IS, lower numbers indicate better performance. The ticks indicate whether a component is used. The acronyms and symbols are defined as:
    HF: high-fidelity, LF1: low-fidelity 1, LF2: low-fidelity 2, LF3: low-fidelity 3, $\mathcal{L}_{IS}$: negatively oriented interval score term in the loss function of \Cref{eqn:loss_function}, PB1: probabilistic Block 1, PO: probabilistic output.}
    \label{tab:ablationStudy}
\end{table}
We summarize the results of the ablation study on the two examples in \Cref{tab:ablationStudy}. For both problems, we observe that using all components minimizes the test MSE and IS. Notably, both of our model's probabilistic components significantly increase the performance: the probabilistic output enables \acronym to not only capture aleatoric uncertainty, but also leverage IS in its loss function. 
Additionally, using a BNN improves \acronymNoSpace's \hf emulation capabilities by preventing overfitting in scarce data regions (since Block 1 is regularized) and by partially disentangling epistemic and aleatoric uncertainties which yields better PIs.

We observe that without a probabilistic output, the IS (and hence the uncertainty quantification accuracy) drops quite significantly (compare V3 to V1 and the base in either of the problems) since the model can no longer account for aleatoric uncertainties. By comparing V1 to the base model in either of the problems in \Cref{tab:ablationStudy} we see that for a model with a probabilistic output the optimal performance is obtained when $\loss_{IS}$ is used in the loss. That is, leveraging the IS in training improves both mean prediction and uncertainty quantification (measured via MSE and IS, respectively).

In both problems, evaluating V1 through V3 against one another indicates that there is a trade-off between MSE and IS. That is, versions that perform well in terms of MSE, do not generally provide the smallest IS. However, when all of these components are included in \acronym (see the base model in \Cref{tab:ablationStudy} for either of the problems), both MSE and IS are reduced. This improvement is due to the fact that the priors and $\loss_{IS}$ effectively regularize the model whose learning capacity is substantially increased by the probabilistic natures of Block 1 and the output. 

The probabilistic fidelity manifold (i.e., output of Block 1) provides an intuitive and visualizable tools to learn the similarity/discrepancy among the sources. Hence, once we fit the base model in each problem, we analyze the learned fidelity manifold to determine the LF source that has the least similarity to the HF source, see \Cref{fig:LS_analyticproblems}(a) and \Cref{fig:LS_realworldproblems}(a). Based on the distances in the fidelity manifold of each problem, we conclude that the third LF source is the least correlated one with the HF source in both cases. We exclude this source and its data from MF modeling and refit the base model to the rest of the data, see version V4 for both problems.

One of the major outputs of \acronym is the learned fidelity manifold which indicates which LF source has the highest discrepancy compared to the HF source. Hence, after training a \acronym and inversely identifying the least accurate LF source, we can build another \acronym while excluding the data from this source.
In the Rational problem, omitting the lowest-fidelity source results in much worse MSE and IS. We explain this observation by noting that this problem has an extremely small number of \hf samples and therefore it is important to judiciously use all available data in training. 
However, in the DNS-ROM problem version V4 achieves the best MSE while \acronym with all components achieves the best IS and second best MSE (compare base to V4 in \Cref{tab:ablationStudy}). We explain this trend by noting that the size of the training data in the DNS-ROM problem is significantly higher than that in the Rational problem. Therefore, omitting a highly inaccurate data source improves mean prediction accuracy for the \hf source in the DNS-ROM problem since the input-output relationships learned by Block 3 for the different sources are more similar. Omitting data from this source also increases the ratio of \hf data available in the unified data set which helps in learning the \hf behavior. However, using all data sources provides \acronym with more information which improves the uncertainty quantification capability and hence a smaller error on IS. 



\subsection{Analytic Problems}
\label{sec: results-analytic}

In this section, we validate our approach against LMGP and existing NN-based technologies for the Rational, Wing-weight, and Borehole examples detailed in \Cref{sec: app-analytic}. These examples cover a wider range of input dimensionality, number of sources, and model form errors (e.g., additive and nonlinear biases). Similar to the previous section, we use MSE and IS on \hf test data as the performance metrics. 
The input space of these three examples does not have categorical features and hence both \acronym and LMGP learn a single manifold. 
We visualize the fidelity manifolds learned by \acronym and LMGP to examine these models' ability in inversely learning the relationships among the data sources (note that the LF sources are \textit{not} ordered based on their accuracy). We highlight that, unlike \acronymNoSpace, the fidelity manifold of LMGP is not probabilistic and hence each data source is encoded with a single point in the manifold. 

\begin{table}[!b]
    \centering
    \begin{tabular}{ccc|cc|cc}
        \cline{2-7}
        & \multicolumn{2}{c|}{Rational} & \multicolumn{2}{c|}{Wing-weight} & \multicolumn{2}{c}{Borehole} \\ \hline
        \multicolumn{1}{c|}{Model} & MSE                    & IS      & MSE             & IS             & MSE           & IS           \\ \hline
        \multicolumn{1}{c|}{\acronym}  & $\bm{1.65 \times 10^{-3}}$  & $0.22$   & $59.14$           & $37.09$          & $14.94$         & $19.24$        \\
        \multicolumn{1}{c|}{LMGP}  & $1.70 \times10^{-3}$   & $\bm{0.20}$  & $\bm{37.97}$           & $\bm{29.70}$          & $\bm{12.69}$         & $\bm{17.57}$        \\
        \multicolumn{1}{c|}{FFNN}  & $2.95 \times 10^{-3}$   & -       & $64.23$           & -              & $21.16$         & -            \\
        \multicolumn{1}{c|}{SMF}   & $8.08 \times 10^{-3}$    & -       & $542.74$          & -              & $172.87$        & -            \\ \hline
    \end{tabular}
    \caption{\textbf{Results on the analytic examples for different models:} We test the performance of \acronym against LMGP and existing NN-based technologies for the Rational, Wing-weight and Borehole examples detailed in \Cref{tab:analytic_functions}. The training procedure for \acronymNoSpace, LMGP, FFNN and SMF is discussed in \Cref{sec: meth}, \Cref{sec: back-LMGP}, \Cref{sec: app-FFNN} and, \Cref{sec: app-SMF} respectively. We report the MSE and IS on unseen HF data.}
    \label{tab:results_analytic}
\end{table}

The results for each approach on each problem are summarized in \Cref{tab:results_analytic} and demonstrate that the probabilistic approaches, \ie LMGP and \acronymNoSpace, significantly outperform the deterministic approaches in all problems. The FFNN approach performs significantly worse than LMGP and sometimes approaches the performance of \acronym in MSE, while the SMF approach shows poor performance for all problems. We explain SMF's poor performance by noting that, as explained in \Cref{sec: app-NN_approaches}, hierarchical MF techniques such as SMF heavily rely on the knowledge of fidelity levels to process the data sources sequentially in the order of increasing accuracy. Since we assume in the problem setup that we only know which source has the highest fidelity and do not know the relative fidelity levels of the \lf sources, the \lf sources are ordered sub-optimally in the SMF approach which leads to a very poor prediction accuracy. The FFNN approach, by contrast, does not rely on the knowledge of fidelity levels and as such performs better than SMF. However, its performance lags behind that of LMGP and \acronym because the architecture is not designed with MF problems in mind.

\begin{figure*}[!h] 
	\centering
	\includegraphics[page=1, width = 1.0\textwidth]{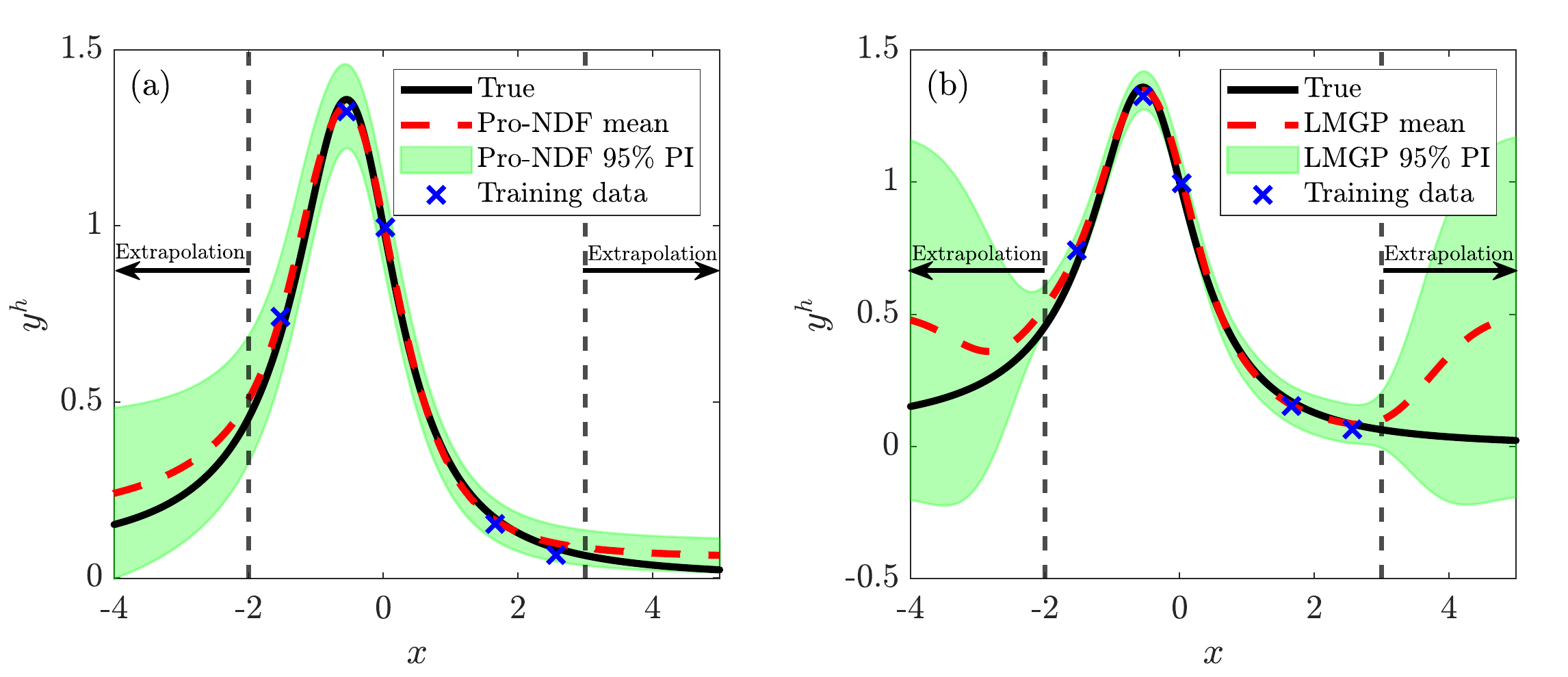}
	\vspace{-0.3cm}
	\caption{\textbf{High-fidelity emulation on the Rational problem}: \acronym and LMGP approaches produce similar results in terms of mean prediction in interpolation. However, LMGP has a narrower $95\%$ PI and reverts to its mean in extrapolation.}
	\label{fig:results_1D_LMGP_novelNN}
\end{figure*}

LMGP, which is considered as our gold standard for MF problems with small data, outperforms \acronym in both MSE and IS for the Wing-weight and Borehole problems, and in IS for the Rational problem. The Rational problem is simultaneously the most data deficient and least complex of the problems examined in this paper: as shown in \Cref{tab:analytic_functions}, there are 4 data sources with only one being especially inaccurate, the input and output are both $1D$, and there are only $5$ training samples provided for the \hf source. 
\acronym and LMGP are well suited to tackle this problem as they both perform well for low-dimensional problems with simple underlying functional forms and well-correlated sources, and as such they have similar performance. \Cref{fig:results_1D_LMGP_novelNN}(a,b) reveals that LMGP captures all of the training points in a narrower $95\%$ PI compared to \acronymNoSpace which explains LMGP's lower IS in \Cref{tab:results_analytic}. However, \acronym shows a better performance for this problem in terms of mean prediction accuracy and it also has a higher degree of agreement with the true function in extrapolation while LMGP reverts to its mean. We therefore conclude that both methods perform on par on the Rational problem.

\begin{figure*}[!t] 
    \centering
    \includegraphics[page=1, width = 1.0\textwidth]{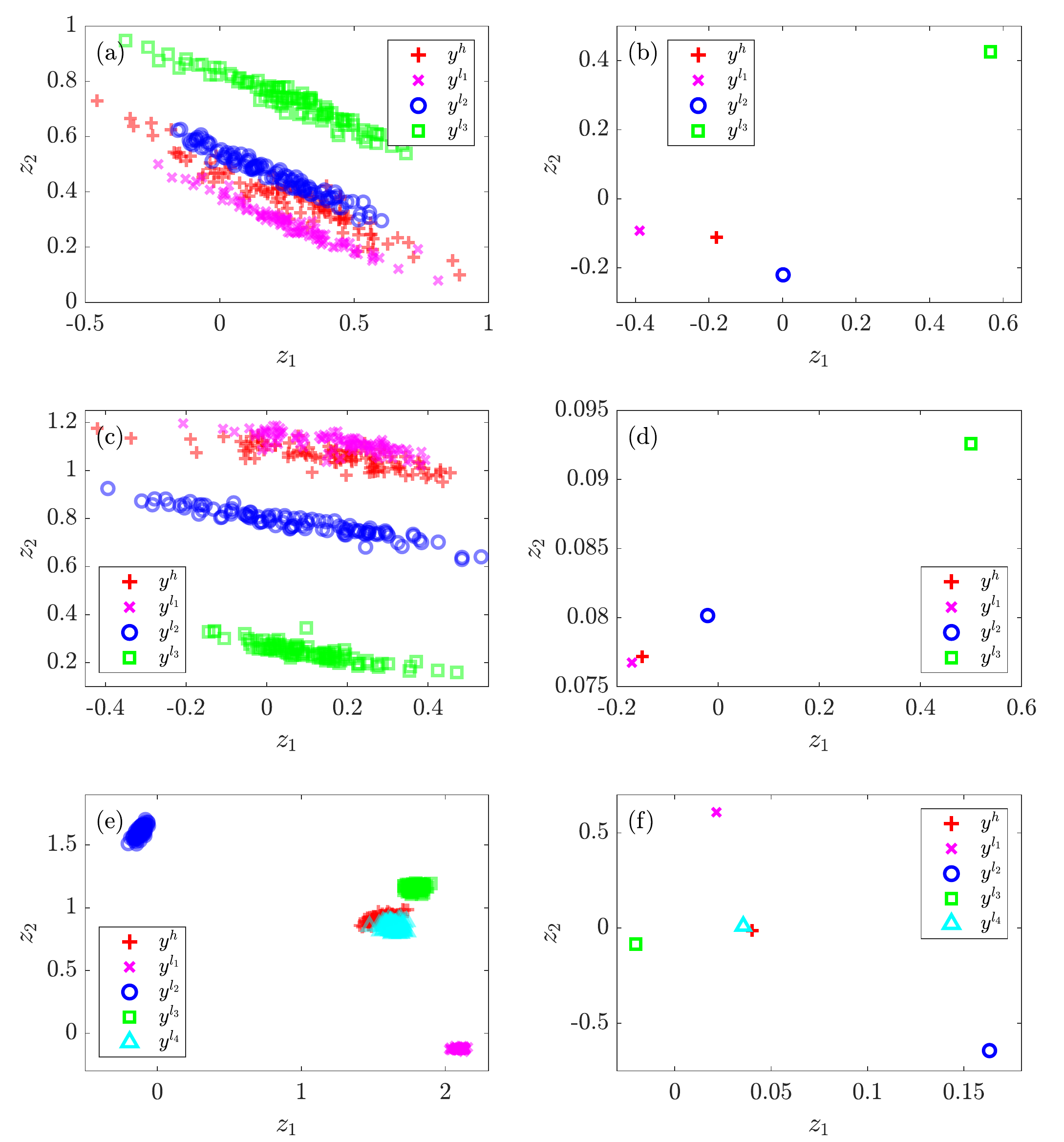}
    \vspace{-0.3cm}
    \caption{\textbf{\acronym and LMGP fidelity manifolds for the analytic problems:} (a) \acronym for Rational problem, (b) LMGP for Rational problem, (c) \acronym for Wing-weight problem, (d) LMGP for Wing-weight problem, (e) \acronym for Borehole problem, and (f) LMGP for Borehole problem.}
    \label{fig:LS_analyticproblems}
\end{figure*}

The learned fidelity manifold of \acronym for the Rational problem is shown in \Cref{fig:LS_analyticproblems}(a) which indicates that the network has inversely learned the true relationship between the data sources as $\yl{1}$ and $\yl{2}$ are encoded close to $\yh$ while $\yl{3}$ is quite far from $\yh$. These relative distances are proportional to the accuracy of the LF sources with respect to the \hf source which are reported in \Cref{tab:analytic_functions}. The fidelity manifold also shows a high spread in the distributions of the realizations for individual sources which indicates either a poor fit to the data or a lack of training samples. In this case, we attribute this spread to lack of data since the performance in IS and MSE is quite good.

The Wing-weight and Borehole problems are both high-dimensional problems with relatively complex underlying functional forms and small amounts of data. LMGP is very well suited to tackle this type of problem\cite{eweis2022data} because the number of its hyperparameters scales much better than NN-based approaches such as \acronym. Accordingly, we observe that LMGP achieves lower MSE and IS for both examples. 

Comparing the performance of \acronym across the two high-dimensional problems, we observe that it performs much better on the Borehole problem. Examining the fidelity manifold learned by \acronym and LMGP for the Wing-weight problem, see \Cref{fig:LS_analyticproblems}(c) and \Cref{fig:LS_analyticproblems}(d), respectively, we see that both approaches accurately determine the relationship between the sources as they agree with the RRMSEs reported in \Cref{tab:analytic_functions}.
Specifically, $\yl{1}$ is closer to $\yh$ than $\yl{2}$, which in turn is closer than $\yl{3}$. Notably, both LMGP and \acronym have the same relative ordering and positioning of the sources, \ie $(1)$ the mean position of all sources lies on an axis, and $(2)$ $\yl{1}$ is in the opposite direction relative to $\yh$ from $\yl{2}$ and $\yl{3}$. This reinforces our earlier assertion in \Cref{sec: meth}: the positions of the sources in the fidelity manifold learned by \acronym reflect \textit{correlations} between the data sources. However, the relative distances between the \lf sources in the latent space found by LMGP more accurately represents the true relationships between the sources because the position for $\yl{3}$ is much more distant from $\yh$ than encoded positions of the other sources. 

In \Cref{fig:LS_analyticproblems}(a) we observe a large spread in the realizations (i.e., the posterior distributions in the fidelity manifold are quite wide) which partially explains the poor\footnote{Poor with respect to LMGP. The performance of \acronym is still much better than the other two NN-based approaches.} performance in this problem. We attribute this performance level to the relative accuracy of the data sources since only one source, $\yl{1}$, is at all accurate with respect to $\yh$ while the other \lf sources are quite inaccurate. LMGP's performance is not inherently hampered by including poorly correlated sources in the data fusion problem \cite{eweis2022data} since its performance, upon successful optimization, is at worse on par with fitting separate GPs to each source. By contrast, since \acronymNoSpace's Block 3 is responsible for learning the relations between all sources and uses weight sharing, including especially inaccurate sources leads to relatively poor performance as shown in \ref{sec: results-ablation}. 

The Borehole problem has five total sources where two \lf sources ($\yl{3}$ and $\yl{4}$) are accurate while two \lf sources ($\yl{1}$ and $\yl{2}$) are quite inaccurate with respect to $\yh$. Since there are more total data available compared to the Wing-weight problem due to the additional data source, and since there are more high-accuracy LF sources, we observe that the spread of the realizations in the fidelity manifold of \acronym is much smaller than in the Wing-weight, see \Cref{fig:LS_analyticproblems}(e). This narrow spread indicates that a good fit has been achieved. We again observe that the relative directions and distances of the \lf sources from $\yh$ are nearly identical in the manifolds of \acronym and LMGP, see \Cref{fig:LS_analyticproblems}(f), and that both methods have correctly identified the relationships between the sources, see \Cref{tab:analytic_functions}. Based on these observations, it is no surprise that \acronym achieves very good performance and nearly matches LMGP in terms of MSE and IS.

\subsection{Real-World Problems}
\label{sec: results-real_world}

In this section, we validate our approach against LMGP and existing NN-based technologies on two engineering applications which are detailed in \Cref{sec: app-real_world}. We again use MSE and IS on \hf test data as our performance metrics and examine the manifolds learned by \acronym and LMGP. In both of these applications, the input space has categorical features (so \acronym and LMGP each build two manifolds) and we do not know the underlying relationships between the data sources.

\begin{table}[!b]
    \centering
    \begin{tabular}{ccc|cc}
        \cline{2-5}
                                   & \multicolumn{2}{c|}{DNS-ROM}                 & \multicolumn{2}{c}{HOIP} \\ \hline
        \multicolumn{1}{c|}{Model} & MSE                   & IS                   & MSE         & IS         \\ \hline
        \multicolumn{1}{c|}{\acronym}  & $\bm{8.99\times 10^{9}}$  & $\bm{4.84\times 10^{5}}$ & $\bm{14.16}$       & $\bm{14.84}$  \\
        \multicolumn{1}{c|}{LMGP}  & $9.66\times 10^{9}$ & $6.60\times 10^{5}$ & $14.33$       & $20.08$      \\
        \multicolumn{1}{c|}{FFNN}  & $1.12\times 10^{10}$ & -                    & $21.55$       & -          \\
        \multicolumn{1}{c|}{SMF}   & $1.81\times 10^{10}$ & -                    & $28.09$       & -          \\ \hline
    \end{tabular}
    \caption{\textbf{Results on the real-world examples for different models:} We test the performance of \acronym against LMGP and existing NN-based technologies for the DNS-ROM and HOIP data sets detailed in \Cref{sec: app-real_world}. We report the MSE and IS on unseen HF data.}
    \label{tab:results_realworld}
\end{table}
\begin{figure*}[!t] 
    \centering
    \includegraphics[page=1, width = 1.0\textwidth]{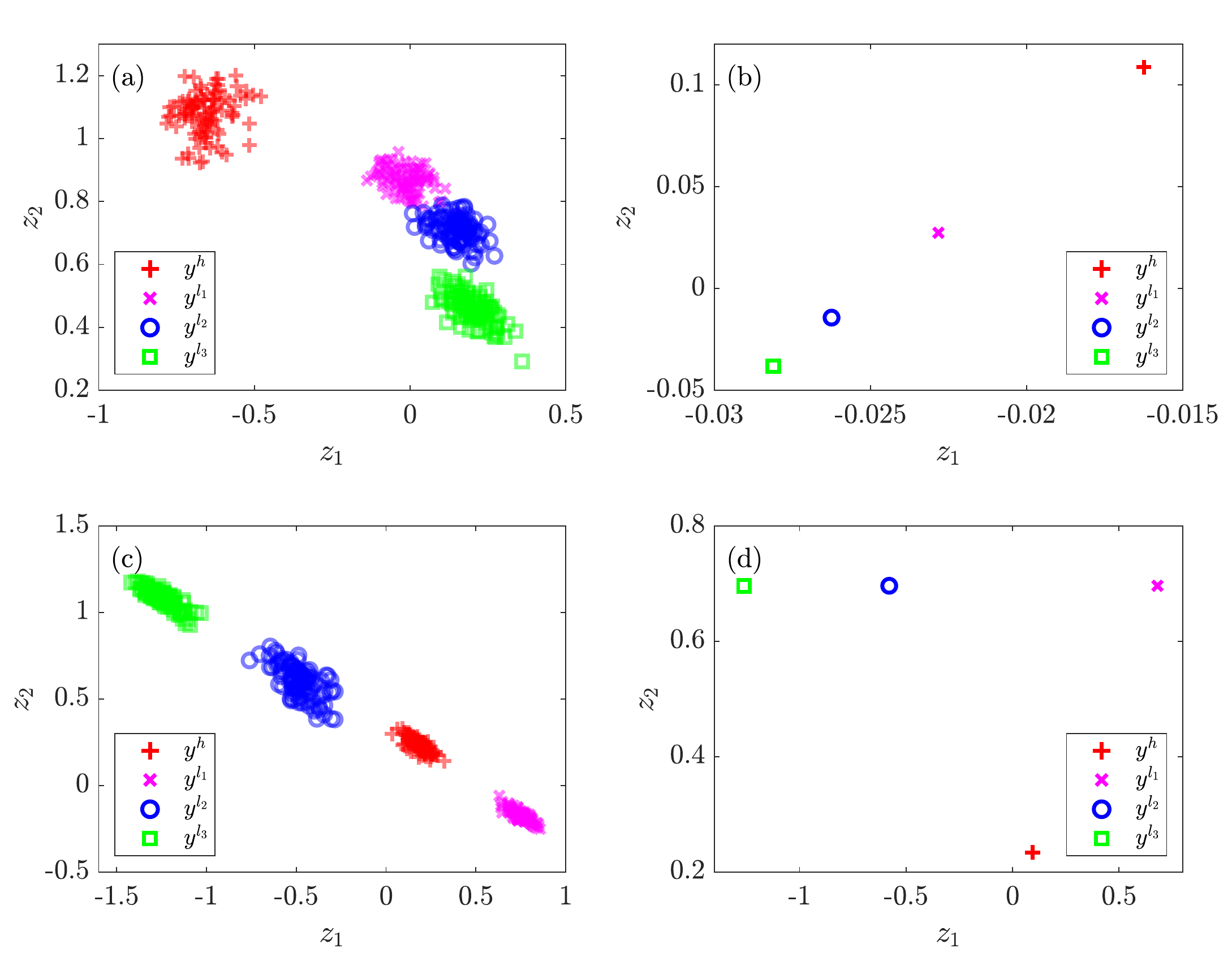}
    \vspace{-0.3cm}
    \caption{\textbf{\acronym and LMGP fidelity manifolds for the real-world problems:} (a) \acronym for DNS-ROM data set, (b) LMGP for DNS-ROM data set, (c) \acronym for HOIP data set, (d) LMGP for HOIP data set.}
    \label{fig:LS_realworldproblems}
\end{figure*}
The results for each approach on each problem are summarized in \Cref{tab:results_realworld} and demonstrate that the probabilistic approaches again significantly outperform the deterministic ones. The FFNN approach performs nearly as well as LMGP and \acronym in the DNS-ROM problem, but lags behing \acronym and LMGP in the HOIP problem in terms of MSE. The SMF approach shows poor performance for both problems for the same reasons provided in \Cref{sec: results-analytic}. 
Notably, \acronym outperforms LMGP for both problems in terms of both metrics which we partially explain by noting that there are much more data are available in these real-world problems compared to the analytical examples of \Cref{sec: app-analytic}. Being an NN-based approach, \acronym scales very well with additional data while the performance of LMGP has diminishing returns and eventually plateaus (recall that the latent map and kernel of LMGP are \textit{not} tuned which contribute to this plateauing performance).  

\begin{figure*}[!t] 
    \centering
    \includegraphics[page=1, width = 1.0\textwidth]{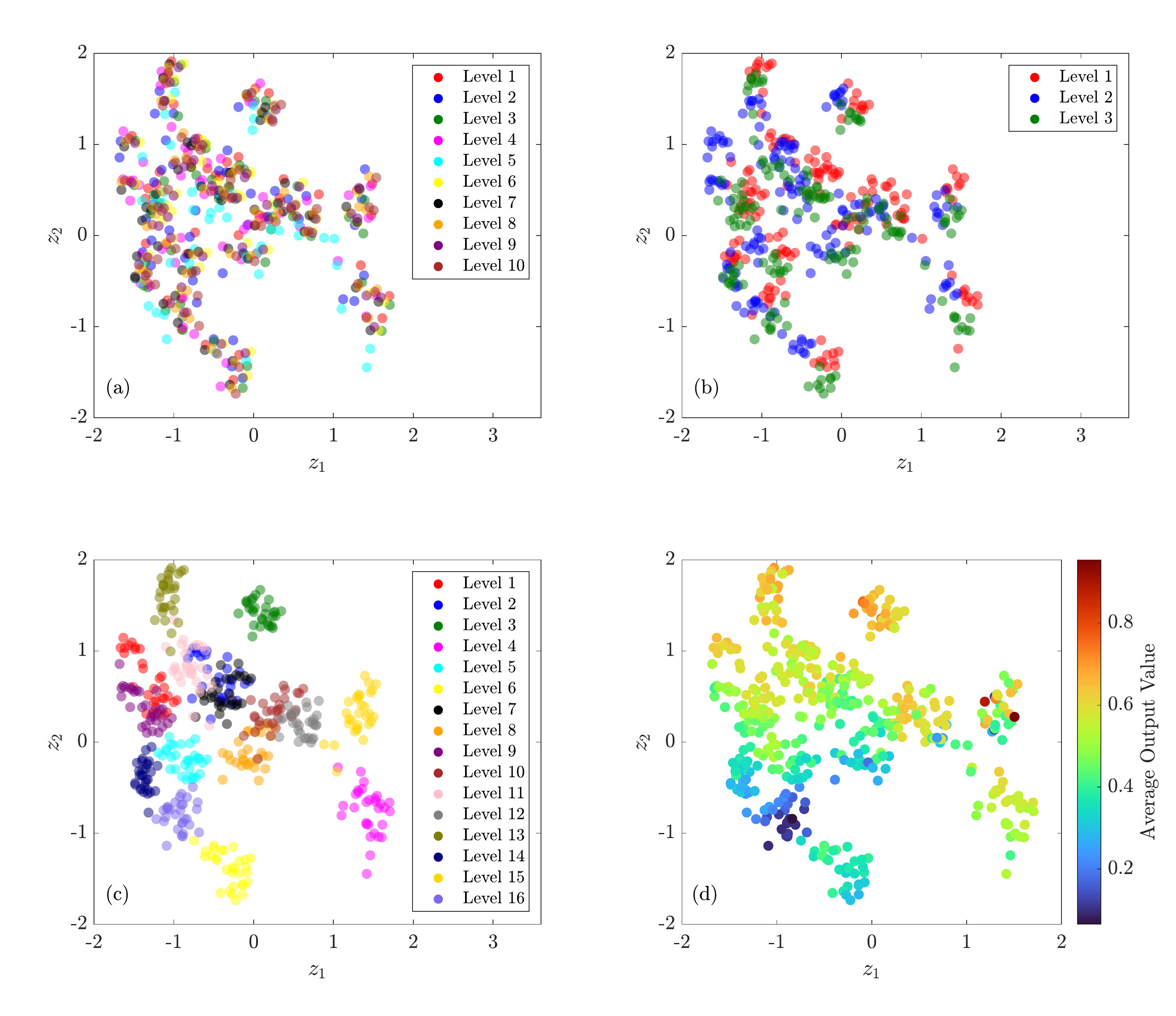}
    \vspace{-0.3cm}
    \caption{\textbf{\acronym categorical manifold for the HOIP problem:} The combination of the categorical variables' levels are color-coded based on: (a) the levels of $t^c_1$, (b) the levels of $t^c_2$, (c) the levels of $t^c_3$, (d) the average output value.}
    \label{fig:LS_HOIP_NNBLM}
\end{figure*}

\begin{figure*}[!t] 
    \centering
    \includegraphics[page=1, width = 1.0\textwidth]{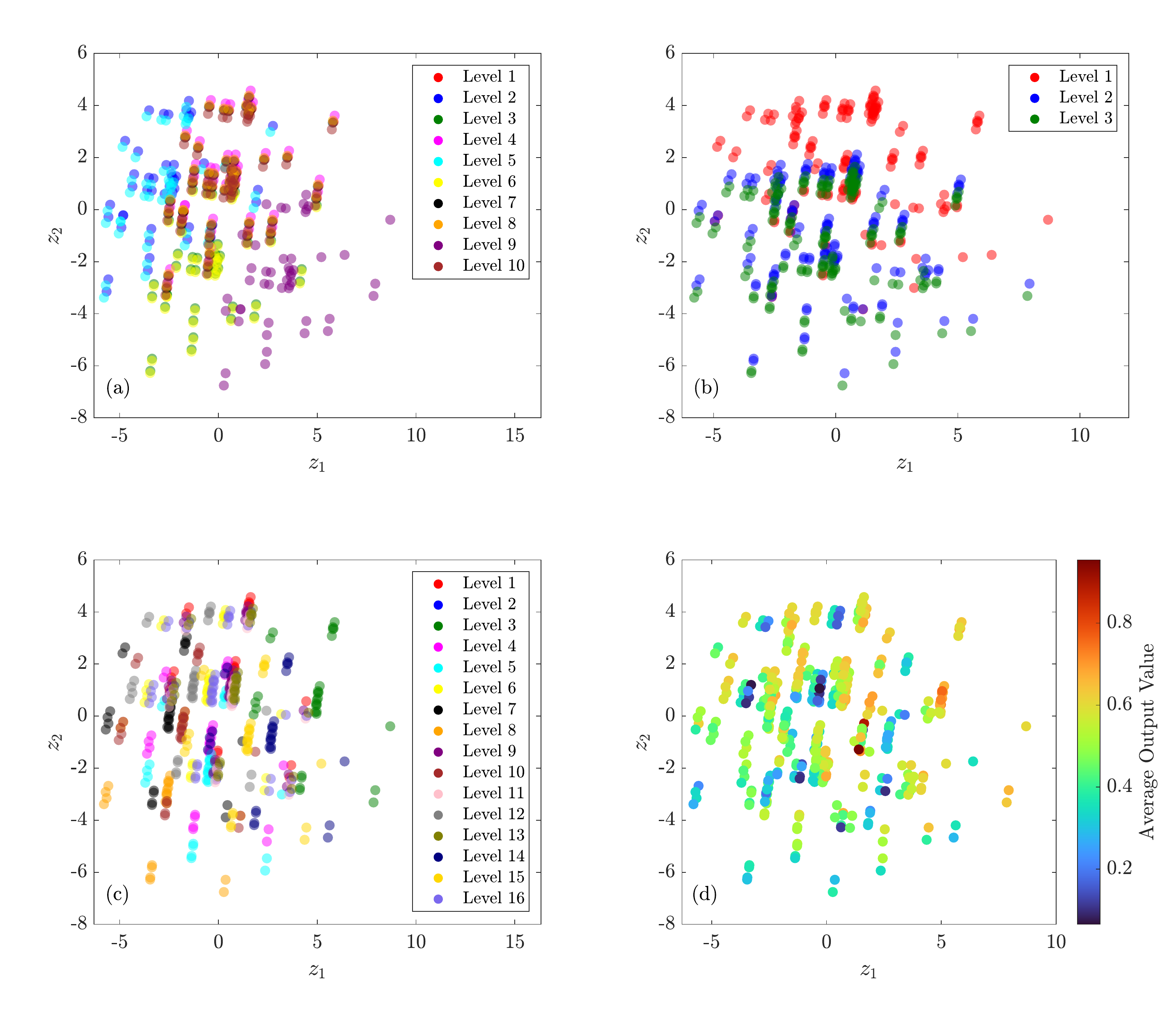}
    \vspace{-0.3cm}
    \caption{\textbf{LMGP categorical manifold for the HOIP problem:} The combination of the categorical variables' levels are color-coded based on: (a) the levels of $t^c_1$, (b) the levels of $t^c_2$, (c) the levels of $t^c_3$, (d) the average output value.}
    \label{fig:LS_HOIP_LMGP}
\end{figure*}

As shown in \Cref{fig:LS_realworldproblems}(a-b), the fidelity manifolds learned by \acronym and LMGP for the DNS-ROM problem are nearly analogous as the relative distances are quite similar. However, LMGP finds all sources to be on the diagonal axis while \acronym learns a more nuanced relationship between the sources, which may contribute to its superior performance. We also observe that the spreads in the individual realizations for each point are fairly tight, which indicates that \acronym is able to learn the relations between the sources reasonably well and, accordingly, provide good performance in terms of MSE and IS.

The HOIP problem has three categorical inputs with $10$, $3$, and $16$ levels and as such \acronym uses two separate latent transformations (one for the data source and the other for the three categorical variables) that correspond to Blocks 1 and 2 in \Cref{fig:NN_architecture}. The learned categorical manifolds for \acronym and LMGP are shown in \Cref{fig:LS_HOIP_NNBLM,fig:LS_HOIP_LMGP} where the $\zbc$ is visualized four times as the combinations of the categorical variables are color-coded based on the levels of each of the three categorical variables and based on the average value of the output \footnote{This average is obtained using the entire data set including both the training and test data}. Since there are no numerical features, the combined inputs are $\combinp = \brackets{\zetab(\ts), \zetab(\tbc)}$ and $\BIIIinp = \brackets{\zbs, \zbc}$ are the inputs to Block 3 of \acronym. Recall that we only use a BNN in Block 1 and as such we show only one realization for the manifold for \acronym that encodes the categorical variables.

\acronym outperforms LMGP in terms of MSE by a small margin and IS by a significant margin for this problem which we attribute to the size of the data sets. \acronym is able to leverage these additional data much more readily than LMGP which only uses simple mapping functions to handle categorical variables $\ts$ and $\tbc$. \acronym also finds fairly tight spreads in the probabilistic fidelity manifold, see \Cref{fig:LS_realworldproblems}(c); indicating that it has high certainty in its outputs and that we should expect good performance. We note that all sources are found to be roughly on one axis and roughly spaced evenly from each other, which may indicate that \acronym has failed to learn the more nuanced relationships between the sources. Equally likely, however, is that the relationships between the sources are simple enough to be represented in this way; since we do not know the underlying functional forms for this problem, we cannot give a definitive answer.

We can also glean some information about the relationships between the categorical variable levels and their impact on the output by examining the corresponding manifolds in \Cref{fig:LS_HOIP_NNBLM,fig:LS_HOIP_LMGP}. \Cref{fig:LS_HOIP_NNBLM}(c) shows that \acronym finds distinct clusters for all $16$ levels of $\tc_3$ which indicates that distinguishing between the levels of $\tc_3$ is important to learning the output. 
Similarly, the levels of $\tc_2$ are distinguishable in \Cref{fig:LS_HOIP_NNBLM}(b) as $\tc_2$ affects the response value. 
By contrast, \Cref{fig:LS_HOIP_NNBLM}(a) shows no apparent trend between the $10$ levels of $\tc_1$ which implies that $\tc_1$ has little effect on the output as \acronym does not learn to distinguish the levels from each other. By contrast, the manifold found by LMGP, shown in \Cref{fig:LS_HOIP_LMGP} shows much less distinct clustering for each of the three categorical variables, which may help explain why it achieves a lower IS than \acronym. Finally, we examine whether the latent positions for the categorical combinations are influenced by the average output value in \Cref{fig:LS_HOIP_NNBLM}(d) and \Cref{fig:LS_HOIP_LMGP}(d). The manifold for \acronym shows a clear trend of the average output value increasing as the latent points move from the bottom-left of the space to the top-right, while for LMGP there is no obvious trend. Based on these manifolds, \acronym shows superior ability to discern relationships between the categorical combinations and between levels of categorical variables.

    \section{Conclusion} \label{sec: conclusion}

In this paper, we introduce \acronym for data fusion under uncertainty. \acronym is based on a multi-block NN where each block is designed to take on specific tasks for MF modeling that arise in typical engineering applications. One of these blocks is probabilistic whose visualizable output can be used to detect LF sources with large model form errors. The final output of \acronym is also probabilistic which enables to not only quantify aleatoric uncertainties, but also leverage strictly proper scoring rules during training. 

We validate each of the key components of \acronym by performing an ablation study on an analytic and a real-world example. 
We also demonstrate that \acronym outperforms other NN-based data fusion approaches by a large margin. Moreover, \acronym performs on par to LMGP in low-dimensional cases with small data sets and slightly lags behind LMGP (a competing GP-based approach) in high-dimensional examples with very small data sets. However, as the size of the training data increases, \acronym scales better than LMGP and provides smaller errors. In these studies, we test the performance on unseen HF data but note that \acronym builds an MF emulator that probabilistically surrogates all the data sources simultaneously. 

A particularly useful output of \acronym is its learnt fidelity manifold which encodes source-wise similarities/discrepancies. While the learnt distances in this manifold do not directly link correlation between the sources, we observe that the fidelity manifold of \acronym and LMGP look quite similar in our studies. Since the fidelity manifold of LMGP is embedded in its kernel and hence indicates the correlations, we believe the fidelity of \acronym also estimates a scaled version of correlation. An added benefit of \acronymNoSpace's fidelity manifold is that it is probabilistic where wide distributions can indicate if \acronym is able to learn the relation between the data sources. Reducing this uncertainty via domain knowledge (especially qualitative information in engineering applications) is a future direction that we plan to investigate. 

The performance of any data fusion approach (including ours) can drop if there are one or more very inaccurate LF sources. With \acronym, the learned fidelity manifold can be used to identify-discard such sources and then retrain \acronym anew. This process can be repeated until all LF sources are encoded close to the HF source in the fidelity manifold. This iterative approach is, however, quite inefficient so we plan to develop an automated mechanism that perhaps leverages the fidelity manifold to adjust the loss function and, in turn, prevent \acronym from learning the highly inaccurate LF sources. 


\section{Acknowledgement}
We appreciate the support from National Science Foundation (award numbers OAC-2211908 and OAC-2103708) and the Early Career Faculty grant from NASA’s Space Technology Research Grants Program (award number 80NSSC21K1809).
    \appendix
\section*{Appendices} \label{sec: app}

\setcounter{equation}{0}
\renewcommand{\theequation}{\thesection-\arabic{equation}}


We provide the formulations of the analytic problems in \Cref{sec: app-analytic}, the background and details of the real-world problems in \Cref{sec: app-real_world}, and the methodology and details of the FFNN and SMF methods in \Cref{sec: app-NN_approaches}.

\section{Table of Analytic Examples}
\label{sec: app-analytic}

\begin{table}[!t]
    \begin{tabular}{|c|c|c|c|c|c|}
    \hline
    \textbf{Name} & \textbf{Source ID} & \textbf{Formulation} & $n$ & $\sigma^{2}$ & \textbf{RRMSE}\\
    \hline
    \multirow{4}{*}{Rational} & $\yh(x)$ & \yhAna & $5$ & $0.001$ & - \Tstrut\Bstrut\\
    \cline{2-6}
     & $\yl{1}(x)$ & \ylIAna & $30$ & $0.001$ & $0.23$ \Tstrut\Bstrut\\
    \cline{2-6}
     & $\yl{2}(x)$ & \ylIIAna & $30$ & $0.001$ & $0.15$ \Tstrut\Bstrut\\
    \cline{2-6}
     & $\yl{3}(x)$ & \ylIIIAna & $30$ & $0.001$ & $0.73$ \Tstrut\Bstrut\\
    \hline
    \multirow{8}{*}{Wing Weight} & \tworow{$\yh(\xb)$} & \yhWWI & \tworow{$15$} & \tworow{$25$} & \tworow{-} \TstrutL\\
     & & \yhWWII & & &  \BstrutL\\
    \cline{2-6}
     & \tworow{$\yl{1}(\xb)$} & \ylIWWI & \tworow{$50$} & \tworow{$25$} & \tworow{$0.20$} \TstrutL\\
     & & \ylIWWII & & &  \BstrutL\\
    \cline{2-6}
     & \tworow{$\yl{2}(\xb)$} & \ylIIWWI & \tworow{$50$} & \tworow{$25$} & \tworow{$1.14$} \TstrutL\\
     & & \ylIIWWII & & &  \BstrutL\\
    \cline{2-6}
     & \tworow{$\yl{3}(\xb)$} & \ylIIIWWI & \tworow{$50$} & \tworow{$25$} & \tworow{$5.75$} \TstrutL\\
     & & \ylIIIWWII & & &  \BstrutL\\
    \hline
    \multirow{5}{*}{Borehole} & $\yh(\xb)$ & \yhBH & $15$ & $6.25$ & - \Tstrut\Bstrut\\
    \cline{2-6}
     & $\yl{1}(\xb)$ & \ylIBH & $50$ & $6.25$ & $3.67$ \Tstrut\Bstrut\\
    \cline{2-6}
     & $\yl{2}(\xb)$ & \ylIIBH & $50$ & $6.25$ & $3.73$ \Tstrut\Bstrut\\
    \cline{2-6}
     & $\yl{3}(\xb)$ & \ylIIIBH & $50$ & $6.25$ & $0.38$ \Tstrut\Bstrut\\
     \cline{2-6}
     & $\yl{4}(\xb)$ & \ylIVBH & $50$ & $6.25$ & $0.19$ \Tstrut\Bstrut\\
    \hline
    \end{tabular}
    \caption{\textbf{Table of analytic functions:} The analytic examples have different input dimensionality, number of sources, and forms of model error. $n$ denotes the number of samples, $\sigma^2$ is the variance of the noise, and RRMSE is the relative root mean squared error of an LF source with respect to an HF source, see \Cref{eqn:RRMSE}.}
    \label{tab:analytic_functions}
\end{table}

\Cref{tab:analytic_functions} details the analytic functions used for the examples covered in \Cref{sec: results}. For each multi-fidelity problem, we calculate the accuracy of each \lf source with respect to the \hf source via relative root mean squared error (RRMSE):
\begin{equation}
    \text{RRMSE} = \sqrt{\frac{(\ylb{}-\yhb)^{T}(\ylb{}-\yhb)}
    {10000\times \text{var}(\yhb)}}
    \label{eqn:RRMSE}
\end{equation}
\noindent where $\ylb{}$ and $\yhb$ are $10000\times 1$ arrays of outputs sampled randomly via Sobol sequence from the \lf and \hf sources, respectively. We use the same sample locations and outputs as our test data when evaluating MSE and IS in \Cref{sec: results-ablation,sec: results-analytic}.

\section{Background on Real-World Examples} \label{sec: app-real_world}

In the DNS-ROM problem, the goal is to predict the toughness of a multiscale metallic component with spatially varying porosity by combining four sources of data: $(1)$ high-fidelity: direct numerical simulations (DNS) and $(2,3,4)$ low-fidelity: a reduced-order model (ROM) with three different number of clusters $(800, 1600, 3200)$ which balance accuracy against computational costs. The data sets have six numerical inputs that include pore volume fraction, number of pores, pore aspect ratio, average nearest neighbor distance among the pores, evolutionary rate parameter, and critical effective plastic strain (the last two inputs govern the damage response of the material under load). The more clusters are used in the ROM, the more similar are its results compared to those of DNS at the expense of a higher computational burden. The data set contains $\nh = 70, \nl{1} = 110, \nl{2} = 170, \nl{3} = 250$ samples. We use $80\%$ of the available samples for each source for training and $20\%$ for testing. For further details on this data set, we refer the reader to \cite{deng2022data}.

In the HOIP problem, the goal is to predict the inter-molecular binding energy in hybrid organic-inorganic perovskite (HOIP) crystals. The data set has three categorical inputs with $l_{1} = 10$, $l_{2}=3$, and $l_{3}=16$ levels which correspond to the elements present in each crystal. There are one HF and three LF data sets with unknown levels of fidelity and $n_{h} = 480, \nl{1} = 480, \nl{2} = 179, \nl{3} = 240$. We use $90\%$ of the available samples for each source for training and $10\%$ for testing.

\section{Other Multi-Fidelity NN-Based Approaches}
\label{sec: app-NN_approaches}

\subsection{Feedforward Neural Networks} \label{sec: app-FFNN}

\begin{figure*}[!t] 
    \centering
    \includegraphics[page=1, width = 0.68\textwidth]{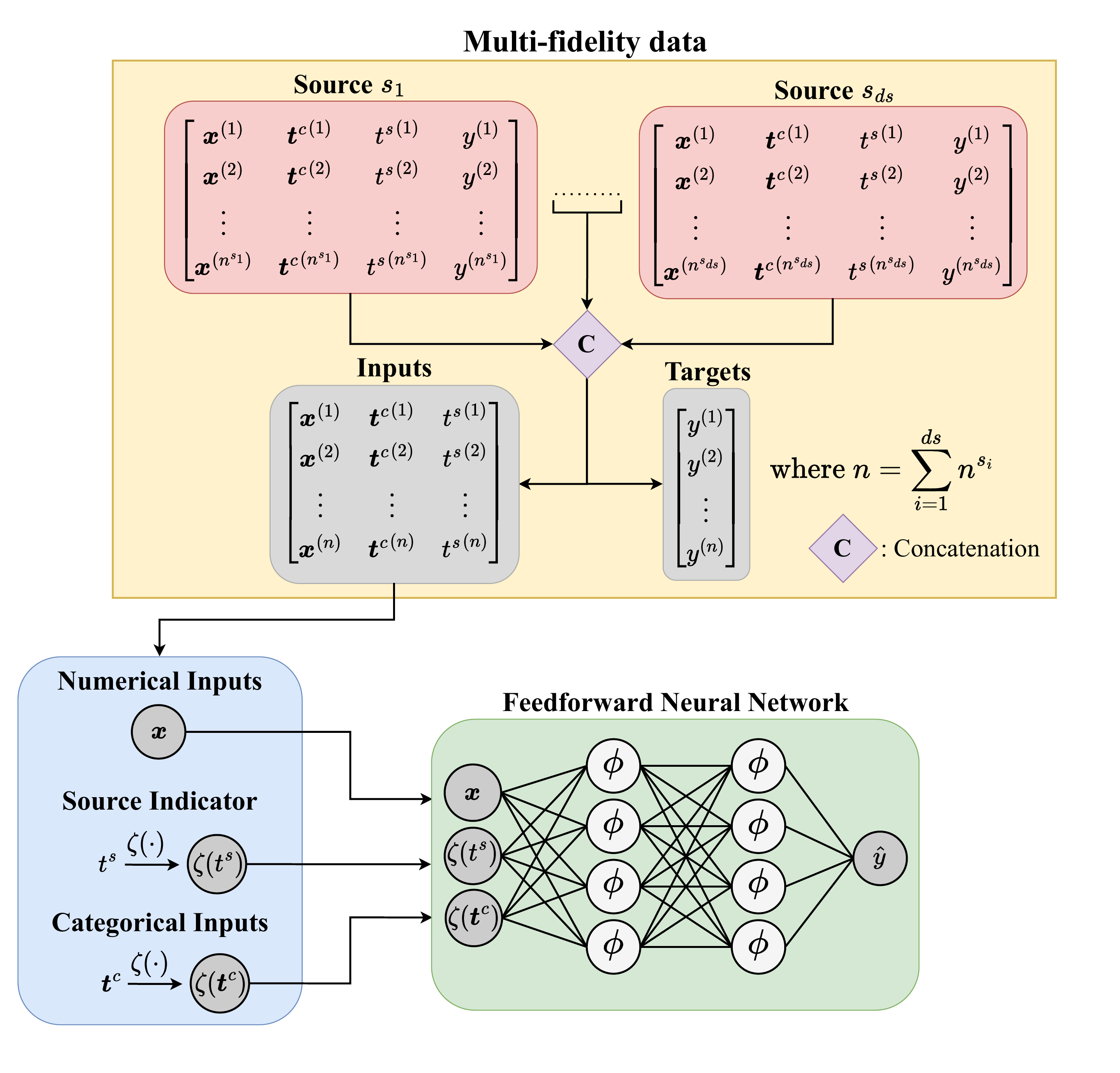}
    \vspace{-0.3cm}
    \caption{\textbf{FFNN for multi-fidelity modeling:} As \acronym, this approach allows to use an arbitrary number of sources by appending a source indicator variable to each data set and concatenating them. The FFNN maps the numerical inputs $\xb$, a priori representation of the source indicator $\zetab(\ts)$, and categorical inputs $\zetab(\tbc)$ to the output.}
    \label{fig:FFNN_archi.pdf}
\end{figure*}
As depicted in \Cref{fig:FFNN_archi.pdf}, for MF modeling via an FFNN we simply feed the numerical inputs $\xb$, the prior representation of the source indicator $\zetab(\ts)$ and the prior representation of the categorical inputs $\zetab(\tbc)$ into the FFNN to produce the output.
This approach has two clear disadvantages with respect to \acronymNoSpace: $(1)$ it does not provide a tool such as the fidelity manifold of \acronym that provides a direct visualization of the correlation between the data sources, and $(2)$ it has a fully deterministic setting which does not enable uncertainty quantification and thus using a loss function based on proper scoring rules. In particular, we use the following loss function for training the FFNN:
\begin{align}
    \mathcal{L} = \mathcal{L}_{MSE} + \beta \mathcal{L}_{2}
    \label{eqn:loss_functionFFNN}
\end{align}
\noindent where $\loss_{MSE}$ is the mean squared error of the predictions and $\loss_2$ is $L2$ regularization: 
\begin{align}
    \loss_{MSE} &= \frac{1}{N}\sum^{N}_{i=1} (\yi - \yhati)^2 \label{eqn:lossMSE}\\
    \loss_{2} &= |\thetab|^2 \label{eqn:lossL2_2}
\vspace{-0.1cm}
\end{align}
We employ Adam as the optimizer and use RayTune \cite{liaw2018tune} and Hyperopt with five-fold cross-validation to find the optimum architecture and hyperparameters which include the learning rate, regularization parameter $\beta$, and batch size $N$. For further details on implementation, please see \href{https://gitlab.com/TammerUCI/pro-ndf}{our GitLab repository}.

\subsection{Sequential Multi-Fidelity Networks} \label{sec: app-SMF}
Unlike the other methods presented in this paper, multi-fidelity modeling via SMF requires training a separate sorrgate for each data source. As depicted in \Cref{fig:SMF_architecture}, individual FFNNs are trained for each source in the sequence that ends with the \hf source. After a sorrugate is trained for a data source, its outputs are used to augment the inputs of the next model in the sequence and hence the resulting input-output relationships are:

\begin{figure*}[!t] 
    \centering
    \includegraphics[page=1, width = 1.0\textwidth]{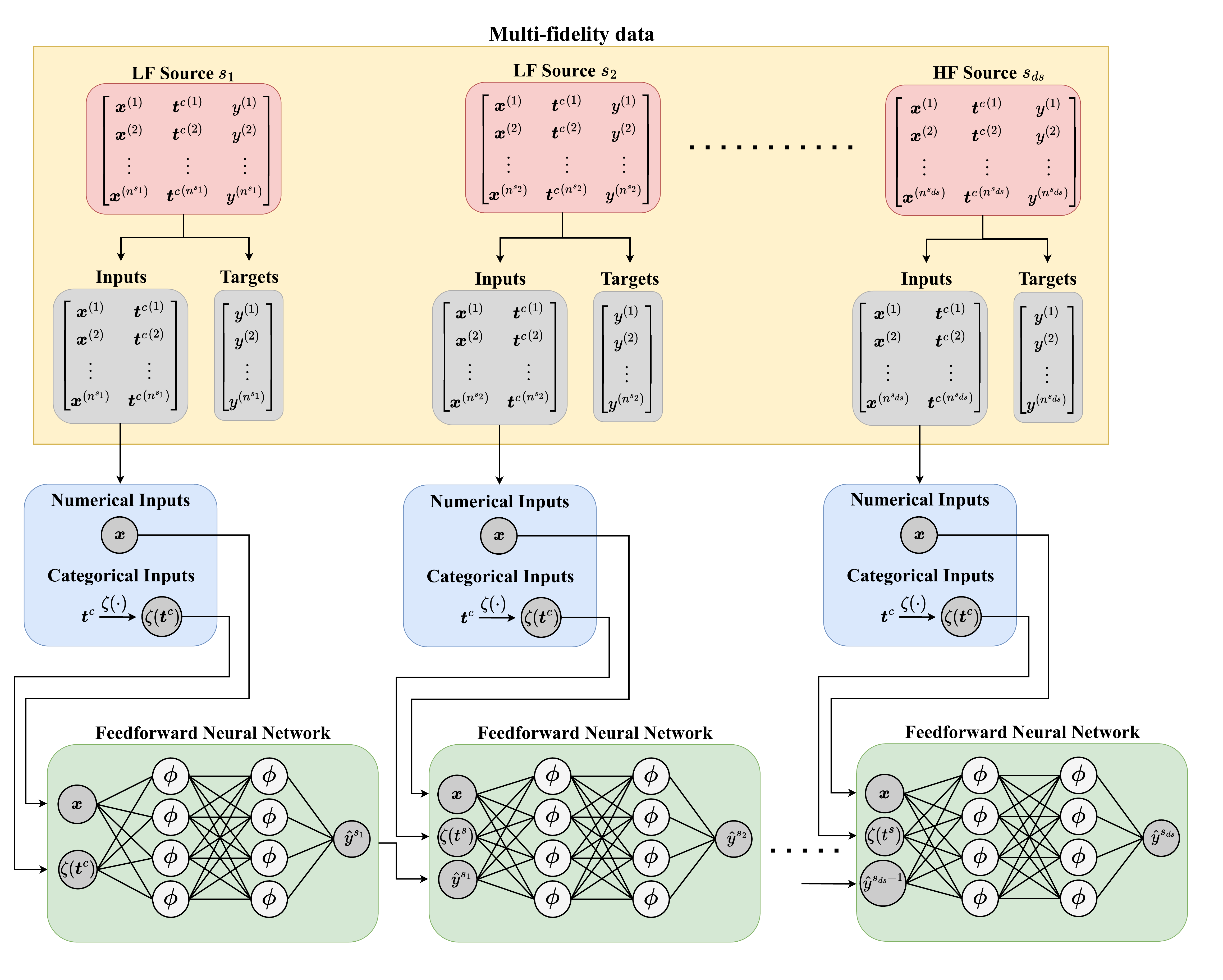}
    \vspace{-0.3cm}
    \caption{\textbf{Sequential Multi-fidelity (SMF) Networks:} SMF is a hierarchical approach that relies on sequentially training a model (e.g., an FFNN) for each data source in an ascending order based on the fidelities. The inputs of a model are augmented with the outputs of the previous one until reaching the model of the HF source.}
    \label{fig:SMF_architecture}
\end{figure*}

\begin{align}
    \yshat{1} &= \fshat{1}\parens{\combinps{1}} \nonumber \\
    \yshat{2} &= \fshat{2}\parens{\combinps{2}, \yshat{1}(\combinps{2})} \nonumber \\
    &\cdots \nonumber \\
    \yshat{ds} &= \fshat{ds}\parens{\combinps{ds}, \yshat{ds-1}(\combinps{ds})}
    \label{eqn:SMF}
\end{align}
\noindent  where $\yshat{i}$ is the output of the FFNN , $\fshat{i}$ is the mapping defined by the FFNN, $\combinps{i}$ is the combined numeric and categorical input $\combinp = \brackets{\xb, \zetab(\tbc)}$, and $i$ denotes the data source with $i=ds$ being the \hf source. Each individual FFNN employs the same loss function and optimizer as in the FFNN method presented in \Cref{sec: app-FFNN}.

Unlike the other three MF methods we study in this paper, the SMF approach is highly sensitive to the ordering of the data sources in the sequence. In the case that the fidelity levels are known, they are assigned in the order of increasing fidelity, \ie source 1 is the least accurate \lf source while source $ds-1$ is the most accurate. With this ordering, the SMF approach leverages the entire data set to achieve good \hf prediction accuracy by minimizing the complexity of the mapping learned by each successive FFNN. 
However, in the case that the fidelities are not known, the order of the \lf sources is assigned randomly. In this case, the mappings of the successive FFNNs no longer monotonically approaches that of the \hf function, and the SMF approach is unable to properly leverage the additional \lf data. In this paper, we assume that the fidelity levels are unknown and therefore assign the data source ordering randomly when using SMF.

Similar to the FFNN approach, the SMF approach does not provide a latent mapping and is entirely deterministic. Like all hierarchical approaches, it also requires knowledge of fidelity levels for good performance. These factors lead to a marked disadvantage in the context of the problems examined in this paper, and we therefore expect the SMF method to perform poorly.

We use RayTune and Hyperopt with five-fold cross-validation to find the optimum architecture and hyperparameters for \textit{each} FFNN in the SMF method. Namely, we tune the learning rate, regularization parameter $\beta$, and batch size N. We also tune an additional parameter that determines whether to use the numeric and categorical inputs $\combinp$ in the final FFNN, since the mapping may be simple enough to learn from just the previous FFNN outputs in the case that the last \lf source is highly accurate. For further details on implementation, please see \href{https://gitlab.com/TammerUCI/pro-ndf}{our GitLab repository}.

    \newpage
    \printbibliography

\end{document}